\documentclass[10pt,journal,compsoc]{IEEEtran}
\ifCLASSOPTIONcompsoc
  \usepackage{cite}
\else
  \usepackage{cite}
\fi
\ifCLASSINFOpdf
\else
\fi
\usepackage{amsmath}

\DeclareMathOperator*{\argmin}{arg\,min}
\usepackage[linesnumbered,ruled]{algorithm2e}

\SetCommentSty{mycommfont}
\usepackage{amsmath,amssymb,amsfonts}
\usepackage[caption=false]{subfig}
\usepackage{threeparttable}
\usepackage{multirow}
\usepackage{tabularx}
\usepackage{cite}
\usepackage{tikz}
\usepackage{threeparttable}

\newcommand{\ineq}[1]{\footnotesize$#1$\normalsize}{}
\newcommand{\fix}[1]{\textcolor{black}{#1}}
\newcommand{\fixnd}[1]{\textcolor{black}{#1}}
\newcommand{\fixa}[1]{\textcolor{black}{#1}}
\newcommand{\mr}[1]{\textcolor{black}{#1}}
\newcommand{\mrr}[1]{\textcolor{black}{#1}}
\newcommand{\tech}{eSpine}{}
\newcommand{\prior}{SpiNeMap}{}
\newcommand{\priorpp}{SpiNeMap++}{}
\usepackage{enumitem}

\begin{document}
\bstctlcite{IEEEexample:BSTcontrol}

\title{\huge{Improving Endurance \fix{and Retention} of Memristive Synapses via Technology-Aware Neuromorphic Compilers}}
\title{\huge{Improving Endurance of Memristive Synapses via Technology-Aware Neuron and Synapse Mapping}}
\title{Endurance-Aware Mapping of Spiking Neural Networks to Neuromorphic Hardware}

\author{Twisha~Titirsha,
        Shihao~Song,
        Anup~Das,
        Jeffrey~Krichmar,
        Nikil~Dutt,\\
        Nagarajan~Kandasamy,~and
        Francky~Catthoor
\IEEEcompsocitemizethanks{\IEEEcompsocthanksitem T. Titirsha, S. Song, A. Das, and N. Kandasamy are with the Department
of Electrical and Computer Engineering, Drexel University,
PA, 19147.\protect\\

E-mail: \{tt624,shihao.song,anup.das,nk78\}@drexel.edu
\IEEEcompsocthanksitem N. Dutt and J. Krichmar are with the Department
of Computer Science, University of California, Irvine, CA, USA.
\IEEEcompsocthanksitem F. Catthoor is with Imec, Belgium and KU Leuven, Belgium.}
\thanks{Manuscript received Month DD, Year; revised Month DD, Year.}}

\markboth{IEEE Transactions on Parallel and Distributed Systems,~Vol.~XX, No.~X, Month~Year}%
{Titirsha \MakeLowercase{\textit{et al.}}: Endurance-Aware Mapping of Spiking Neural Networks to Neuromorphic Hardware}


\IEEEtitleabstractindextext{%
\begin{abstract}
Neuromorphic computing systems are embracing memristors to implement high density and low power synaptic storage as crossbar arrays in hardware. These systems are energy efficient in executing 
Spiking Neural Networks (SNNs).
We observe that long bitlines and wordlines in a memristive crossbar are a major source of parasitic voltage drops, which create 
current asymmetry.
Through circuit simulations, we show the significant endurance variation that results from this asymmetry.
Therefore, if the critical memristors (ones with lower endurance) are overutilized, they may lead to a reduction of the crossbar's lifetime.
We propose \tech{}, a novel technique to improve lifetime by incorporating the endurance variation within each crossbar in mapping machine learning workloads, ensuring that synapses with higher activation are always implemented on memristors with higher endurance, and vice versa. \tech{} works in two steps. First, it uses the Kernighan-Lin Graph Partitioning algorithm to partition a workload into clusters of neurons and synapses, where each cluster can fit in a crossbar. Second, it uses an instance of Particle Swarm Optimization (PSO) to map clusters to tiles, where the placement of synapses of a cluster to memristors of a crossbar is performed 
by analyzing their activation within the workload.
\mr{
We evaluate \tech{} for a state-of-the-art neuromorphic hardware model with phase-change memory (PCM)-based memristors. Using 10 SNN workloads, 
we
demonstrate a significant improvement in the effective lifetime.
}
\end{abstract}

\begin{IEEEkeywords}
Neuromorphic Computing, Spiking Neural Networks (SNNs), Non-Volatile Memory (NVM), Memristor, Endurance.
\end{IEEEkeywords}}

\maketitle

\IEEEdisplaynontitleabstractindextext

\IEEEpeerreviewmaketitle

\IEEEraisesectionheading{\section{Introduction}\label{sec:introduction}}
\IEEEPARstart{S}{piking} Neural Networks (SNNs) are machine learning approaches designed using spike-based computations and bio-inspired learning algorithms~\cite{maass1997networks}. Neurons in an SNN communicate information by sending spikes to other neurons, via synapses. SNN-based applications are typically executed on event-driven neuromorphic hardware such as DYNAP-SE~\cite{dynapse}, TrueNorth~\cite{truenorth}, and Loihi~\cite{loihi}.
These
hardware platforms are designed as tile-based architectures with a shared interconnect for communication~\cite{balaji2019exploration,balaji2019design,catthoor2018very} (see Fig.~\ref{fig:architecture}a). A tile consists of a crossbar for mapping neurons and synapses of an application. Recently, memristors 
such as Phase-Change Memory (PCM) and Oxide-based Resistive RAM (OxRRAM) 
are used to implement high-density and low-power synaptic storage in each crossbar~\cite{Mallik2017,Burr2017,wijesinghe2018all,hu2014memristor,wan202033,wan202033VLSIT}.

As the complexity of machine learning models increases, mapping an SNN to a neuromorphic hardware is becoming increasingly challenging. 
Existing SNN-mapping approaches have mostly focused on improving performance and energy \cite{balaji2020compiling,dfsynthesizer,spinemap,psopart,twisha_thermal,twisha_endurance,balaji2020ESL,rtmJSPS,das2018dataflow,balajiISVLSI}, and reducing circuit aging~\cite{reneu,frameworkCAL,NeuromorphicLR}. Unfortunately, memristors have limited endurance, ranging from
\ineq{10^5} (for Flash) to \ineq{10^{10}} (for OxRRAM), with PCM somewhere in between
(\ineq{\approx 10^{7}}). We focus on endurance issues in a memristive crossbar of a neuromorphic hardware and propose an intelligent solution to mitigate them.

We analyze the internal architecture of a memristive crossbar (see Fig.~\ref{fig:parasitics}) and observe that parasitic components on horizontal and vertical wires of a crossbar are a major source of parasitic voltage drops in the crossbar.
Using detailed circuit simulations at different process (P), voltage (V), and temperature (T) corners, 
\mr{
we show that these voltage drops 
create
current variations in the crossbar.
For the same spike voltage, 
current on the shortest path is significantly higher than the current on the longest path in the crossbar, where the length of a current path is measured in terms of its number of parasitic components.
}
\fixa{
These current variations create
asymmetry in the self-heating temperature of memristive cells during their weight updates, e.g., during model training and continuous online learning~\cite{chen2018lifelong}, which directly influences their endurance.} 

\mr{
The endurance variability in a memristive crossbar becomes more pronounced with technology scaling and at elevated temperature.
If this is not incorporated when executing a machine learning workload, critical memristors, i.e., those with lower endurance may get overutilized, leading to a reduction in the memristor lifetime.
}

In this work, we formulate the \emph{effective lifetime}, a joint metric incorporating the endurance of a memristor, and its utilization within a workload (see Sec.~\ref{sec:mapping}). Our \textbf{goal} is to maximize the minimum effective lifetime. We achieve this goal by first exploiting technology and circuit-specific characteristics of memristors, and then proposing an endurance-aware \textit{intelligent mapping} of neurons and synapses of a machine learning workload to crossbars of a hardware, ensuring that synapses with higher activation are implemented on memristors with higher endurance, and vice versa.


Endurance balancing (also called \textit{wear leveling}) is previously proposed for classical computing systems with Flash storage, where a virtual address is translated to different physical addresses to balance the wear-out of Flash cells~\cite{qureshi2009enhancing,chang2007efficient,yang2014garbage,liao2014adaptive,li2019wear}. 
Such techniques cannot be used for neuromorphic hardware because once synapses are placed to crossbars they access the same memristors for the entire execution duration. 
\mr{
Therefore, it is necessary to limit the utilization of critical memristors of a neuromorphic hardware during the initial mapping of neurons and synapses.
}

To the best of our knowledge, no prior work has studied the endurance variability problem in neuromorphic hardware with memristive crossbars. To this end, we make the following novel \textbf{contributions} in this paper.
\begin{itemize}
    \item We study the parasitic voltage drops at different P, V, \& T corners through detailed circuit simulations with different crossbar configurations. 
    \item \mr{We use these circuit simulation parameters within a compact endurance model to estimate the endurance of different memristors in a crossbar.}
    \item We integrate this endurance model within a design-space exploration framework, which uses an instance of Particle Swarm Optimization (PSO) to map SNN-based workloads to crossbars of a neuromorphic hardware, maximizing the effective lifetime of memristors.
\end{itemize}

The proposed endurance-aware technique, which we call \tech{}, operates in two steps. First, \tech{} partitions a machine learning workload into clusters of neurons and synapses using the Kernighan-Lin Graph Partitioning algorithm such that, each cluster can be mapped to an individual crossbar of a hardware. The objective is to reduce inter-cluster communication, which lowers the energy consumption. 
\mr{
Second, \tech{} uses PSO to map clusters to tiles, placing synapses of a cluster to memristors of a crossbar in each PSO iteration by analyzing their utilization within the workload. The objective is to maximize the effective lifetime of the memristors in the hardware.
We evaluate \tech{} using 10 SNN-based machine learning workloads on a state-of-the-art neuromorphic hardware model using PCM memristors.
}
Our results demonstrate an average 3.5x improvement of the effective lifetime with 7.5\% higher energy consumption, compared to a state-of-the-art SNN mapping technique that minimizes the energy consumption.

\section{Background}\label{sec:background}
\begin{figure}[h!]
	\centering
	\vspace{-10pt}
	\centerline{\includegraphics[width=0.99\columnwidth]{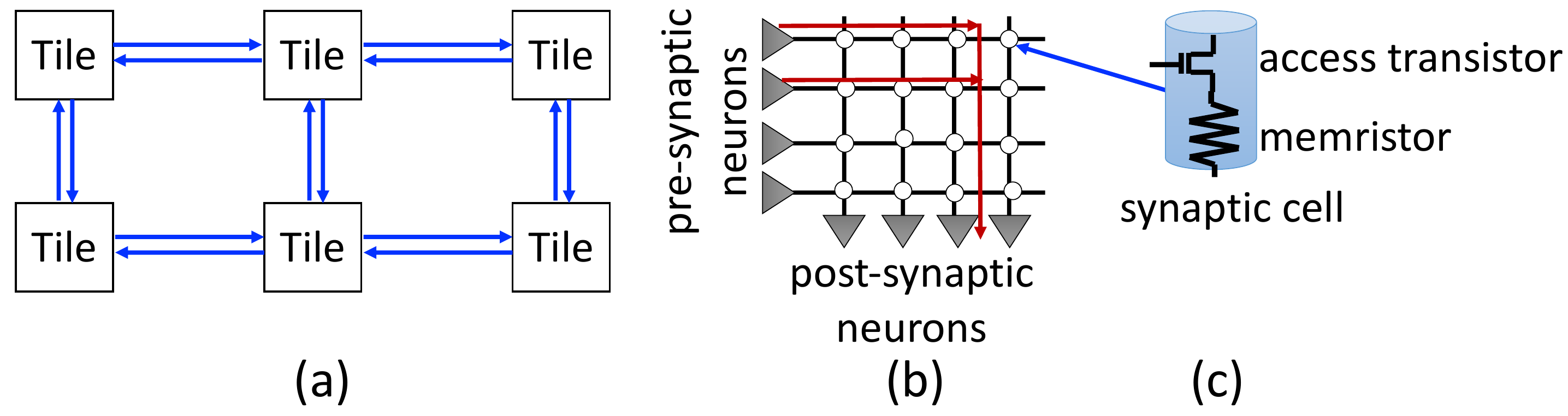}}
	\vspace{-10pt}
	\caption{Neuron and synapse mapping to a tile-based neuromorphic hardware such as DYNAP-SE~\cite{dynapse}.}
	\vspace{-5pt}
	\label{fig:architecture}
\end{figure}

Figure~\ref{fig:architecture}a illustrates a tile-based neuromorphic hardware such as DYNAP-SE~\cite{dynapse}, where each tile consists of a crossbar to map neurons and synapses of an SNN.
A crossbar, shown in Figure~\ref{fig:architecture}b, is 
an organization of row wires called wordlines and column wires called bitlines.
A synaptic cell is connected at a crosspoint, i.e., at the intersection of a row and a column.
Pre-synaptic neurons are mapped along rows and post-synaptic neurons along columns. A \ineq{n\times n} crossbar has \ineq{n} pre-synaptic neurons, \ineq{n} post-synaptic neurons, and \ineq{n^2} synaptic cells at their intersections. Memristive devices such as Phase-Change Memory (PCM)~\cite{Burr2017}, Oxide-based Resistive RAM (OxRRAM)~\cite{Mallik2017}, Ferroelectric RAM (FeRAM)~\cite{jerry2017ferroelectric}, Flash~\cite{bez2003introduction}, and Spin-Transfer Torque Magnetic or Spin-Orbit-Torque RAM (STT- and SoT-MRAM)~\cite{vincent2015spin} can be used to implement a synaptic cell. \footnote{\mr{Beside neuromorphic computing, some of these memristor technologies are also used as main memory in conventional computers to improve performance and energy efficiency~\cite{palp,mneme,datacon,hebe,song2020design}.}} This is illustrated in Figure~\ref{fig:architecture}c, 
where a memristor is represented as a resistance.

We demonstrate \tech{} for PCM-based memristive crossbars. We start by reviewing the internals of a PCM device. The proposed approach can be generalized to other memristors such as OxRRAM and SOT-/STT-MRAM by exploiting their specific structures (see Section~\ref{sec:other_memristors}).

\begin{figure}[h!]
	\begin{center}
		\vspace{-10pt}
		\includegraphics[width=0.79\columnwidth]{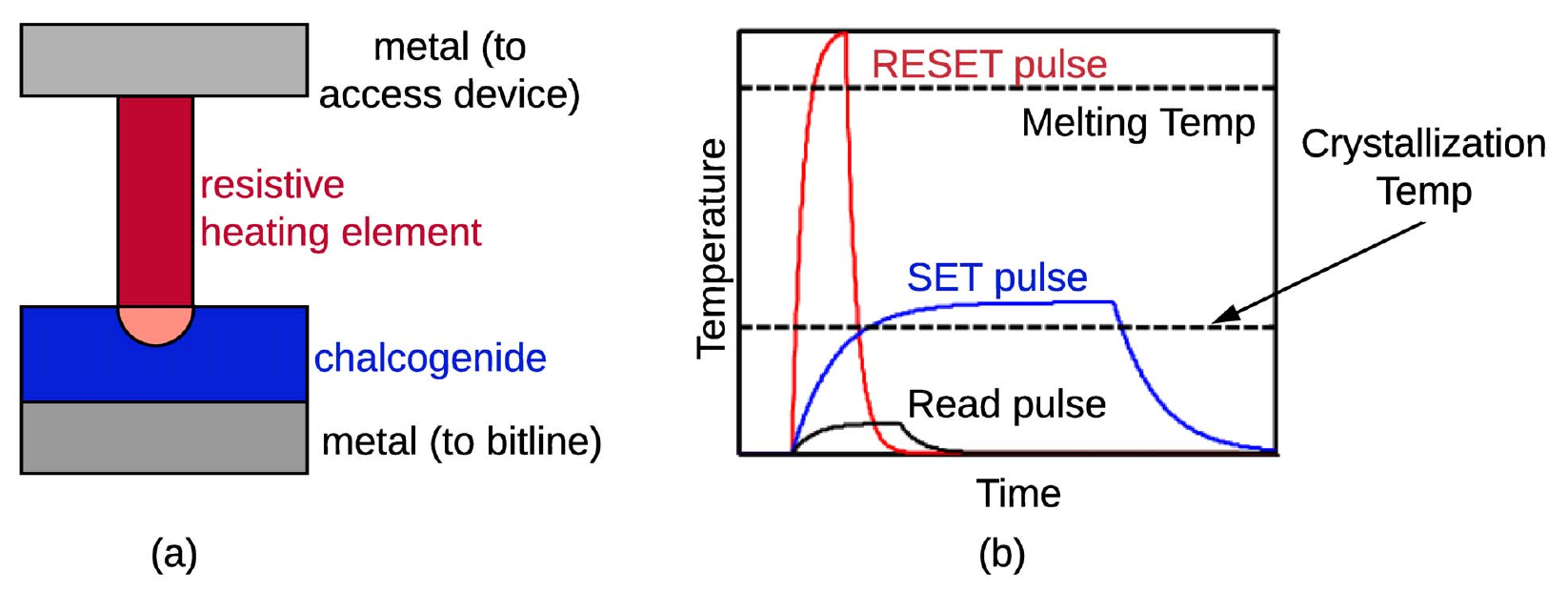}
		\vspace{-10pt}
		\caption{(a) A phase change memory (PCM) cell and (b) current needed to SET, RESET, and read a PCM cell.}
		\label{fig:pcm_memory_cell_integration}
		\vspace{-10pt}
	\end{center}
\end{figure}

Figure \ref{fig:pcm_memory_cell_integration}(a) illustrates how a chalcogenide semiconductor alloy is used to build a PCM cell.
The amorphous phase (logic `0') in this alloy has higher resistance than its crystalline phase (logic `1').
\mr{
When using only these two states, each PCM cell can implement a binary synapse. 
However, with precise control of the crystallization process, a PCM cell can be placed in a partially-crystallized state, in which case, it can implement a multi-bit synapse.
}
Phase changes in a PCM cell are induced by injecting current into resistor-chalcogenide junction and heating the chalcogenide alloy. 
{
Figure \ref{fig:pcm_memory_cell_integration} (b) shows the different current profiles needed to program and read in a PCM device.
}

\section{Analyzing Technology-Specific Current Asymmetry in Memristive Crossbars}\label{sec:current_difference}
\begin{figure}[h!]
	\centering
	\vspace{-10pt}
	\centerline{\includegraphics[width=0.69\columnwidth]{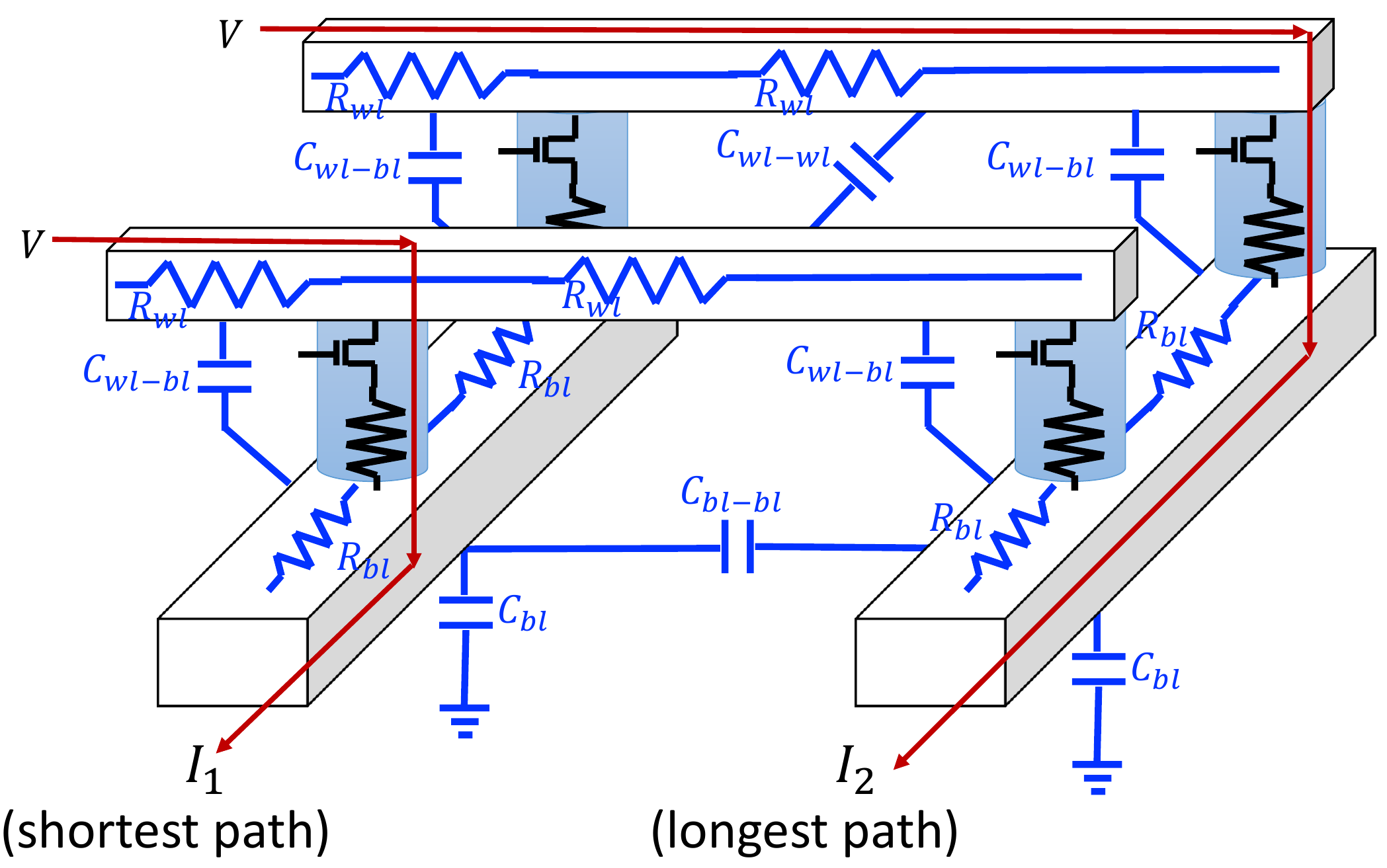}}
	\vspace{-10pt}
	\caption{Parasitcs of bitlines and wordlines in a memristive crossbar.}
	\vspace{-10pt}
	\label{fig:parasitics}
\end{figure}

\mr{
Long bitlines and wordlines in a crossbar are a major source of parasitic voltage drops, introducing asymmetry in current propagating through its different memristors.}
Figure~\ref{fig:parasitics} shows these parasitic components for a 2x2 crossbar. 
We simulate this circuit using LTspice~\cite{kweon2019modelling,junsangsri2012macromodeling} with technology-specific data from predictive technology model (PTM)~\cite{zhao2007predictive}. 
We make the following three key observations.

\emph{\textbf{Observation 1:} The current on the longest path from a pre- to a post-synaptic neuron in a crossbar is lower than the current on its shortest path for the same input spike voltage and the same memristive cell conductance programmed along both these paths.}

\begin{figure}[h!]
	\centering
	\vspace{-10pt}
	\centerline{\includegraphics[width=0.99\columnwidth]{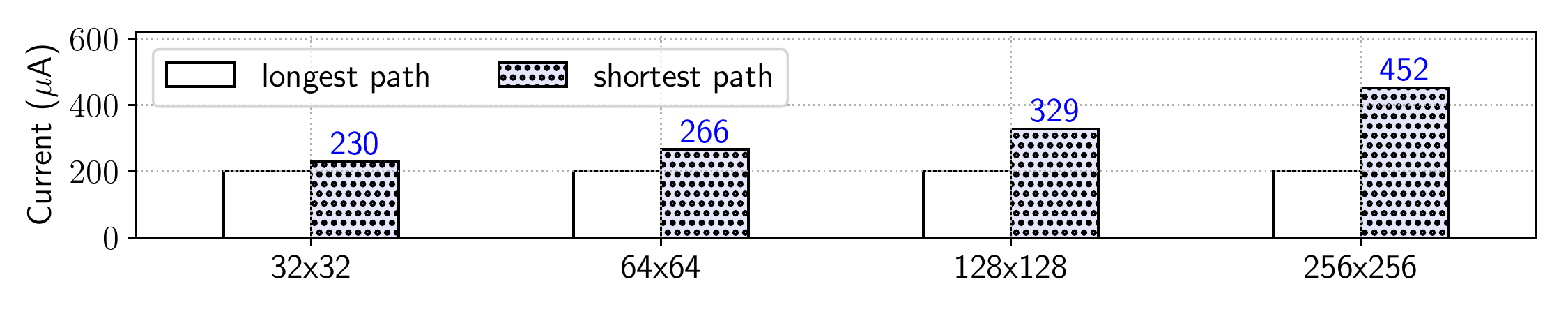}}
	\vspace{-10pt}
	\caption{Difference between current on the shortest and the longest path for different crossbar sizes.}
	\vspace{-10pt}
	\label{fig:current_crossbar_size}
\end{figure}

Figure~\ref{fig:current_crossbar_size} shows the difference between currents on the shortest and longest paths for 32x32, 64x64, 128x128, and 256x256 memristive crossbars at {65nm} process node. The input spike voltage of the pre-synaptic neurons is set to generate \ineq{200\mu A} on ther longest paths. This current value corresponds to the current needed to amorphize the crystalline state of a PCM-based memristor. 

\fixnd{We observe that the current injected into the post-synaptic neuron on the longest path is lower than the current on the shortest path by 13.3\% for 32x32, 25.1\% for 64x64, 39.2\% for 128x128, and 55.8\% for 256x256 crossbar.} 
\mr{
This current difference is because of the higher voltage drop on the longest path, which reduces the current on this path compared to the shortest path for the same amount of spike voltage applied on both these paths.
}
The current difference increases with crossbar size because of the increase in the number of parasitic resistances on the longest current path, which results in larger voltage drops, lowering the current injected into its post-synaptic neuron. Therefore, to achieve the minimum \ineq{200\mu A} current on this path, the input spike voltage must be increased, which increases the current on the shortest path.
This observation can be generalized to all current paths in a memristive crossbar. Current variation in a crossbar may lead to difference in synaptic plasticity behavior and access speed of memristors~\cite{twisha_endurance,fouda2017modeling,woo2018resistive,zhang2020lifetime,wen2019renew}. 
\mr{A circuit-level solution to address the current differences is to add proportional series resistances to the current paths in a crossbar. However, this circuit-level technique can significantly increase the area of a crossbar (\ineq{n^2} series resistances are needed for a \ineq{n}x\ineq{n} crossbar). Additionally, adding series resistances can increase the power consumption of the crossbar.}
\mrr{Although current balancing in a crossbar can be achieved by adjusting the biasing of the crossbar's cells, a critical limitation is that this and other circuit-level solutions do not incorporate the activation of the synaptic cells, which is dependent on the workload being executed on the crossbar. Therefore, some of its cells may get utilized more than others, leading to endurance issues.}
\mr{We propose a system-level solution to exploiting the current and activation differences via intelligent neuron and synapse mapping.}

Current imbalance may not be a critical consideration for smaller crossbar sizes (e.g., for 32x32 or smaller) due to comparable currents along different paths. 
\mr{
However, a neuron is several orders of magnitude larger than a memristor-based synaptic cell~\cite{indiveri2003low}.}
To amortize this large neuron size, neuromorphic engineers implement larger crossbars, subject to a maximum allowable energy consumption. The usual trade-off point is 128x128 crossbars for DYNAP-SE~\cite{dynapse} and 256x256 crossbars for TrueNorth~\cite{truenorth}.


\emph{\textbf{Observation 2:} Current variation in a crossbar becomes significant with technology scaling and at elevated temperatures.}

\begin{figure}[h!]
	\centering
	\vspace{-10pt}
	\centerline{\includegraphics[width=0.99\columnwidth]{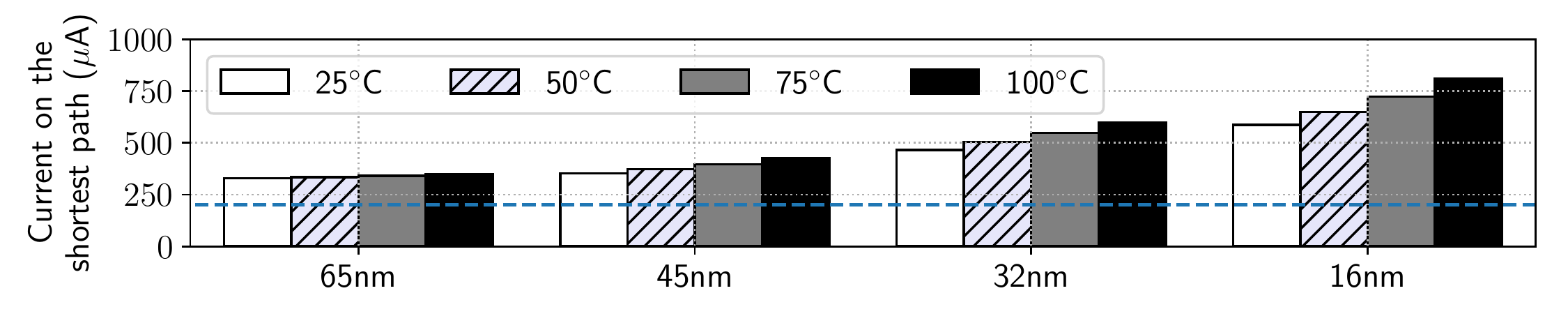}}
	\vspace{-10pt}
	\caption{Current obtained on the shortest path in a 128x128 memristive crossbar at 65nm, 45nm, 32nm, and 16nm technology nodes for 4 ambient temperatures (25$^\circ$C, 50$^\circ$C, 75$^\circ$C, and 100$^\circ$C). The input spike voltage is adjusted to obtain 200$\mu A$ on the longest path.}
	\vspace{-5pt}
	\label{fig:current_128}
\end{figure}

Figure~\ref{fig:current_128} plots the current on the shortest path in a 128x128 memristive crossbar for four process corners (65nm, 45nm, 32nm, and 16nm) and four temperature corners (25$^\circ$C, 50$^\circ$C, 75$^\circ$C, and 100$^\circ$C) with all memristors configured in their crystalline state with a resistance of \ineq{10K\Omega}. \mr{The input spike voltage of the crossbar is set to a value that generates \ineq{200\mu A} on the longest path at each process and temperature corners.} We make two key conclusions.

First, current on the shortest path is higher for smaller process nodes. This is because,
with technology scaling, the value of parasitic resistances along the bitline and wordline of a current path increases~\cite{fouda2017modeling,ciprut2016modeling,son2020signal}. The unit wordline (bitline) parasitic resistance ranges from approximately \ineq{2.5\Omega} (\ineq{1\Omega}) at {65nm} node to  \ineq{10\Omega} (\ineq{3.8\Omega}) at {16nm} node. The value of these unit parasitic resistances are expected to scale further reaching \ineq{\approx 25\Omega} at {5nm} node~\cite{fouda2017modeling}. This increase in the value of unit parasitic resistance increases the voltage drop on the longest path, reducing the current injected into its post-synaptic neuron. Therefore, to obtain a current of \ineq{200\mu A} on the longest path, the input spike voltage must be increased, which increases the current on the shortest path.

Second, current reduces at higher temperature. This is because, the leakage current via the access transistor of each memristor in a crossbar increases at higher temperature, reducing the current injected into the post-synaptic neurons. \mr{To increase the current to \ineq{200\mu A}, the spike voltage is increased, which increases the current on the shortest path.}

Based on the two observations and the endurance formulation in Section~\ref{sec:endurance_model}, we show that higher current through memristors on shorter paths in a memristive crossbar leads to their higher self-heating temperature and correspondingly lower cell endurance, compared to those on the longer current paths in a crossbar. 
Existing SNN mapping approaches such as SpiNeMap~\cite{spinemap}, PyCARL~\cite{pycarl}, DFSynthesizer~\cite{dfsynthesizer}, and SNN Compiler~\cite{ji2018bridge} do not take endurance variation into account when mapping neurons and synapses to a crossbar. Therefore, synapses that are activated frequently may get mapped on memristors with lower cell endurance, lowering their lifetime.


\emph{\textbf{Observation 3:} Synapse activation in a crossbar is specific to the machine learning workload as well as to mapping of neurons and synapses of the workload to the crossbars.} 

\begin{figure}[h!]
	\centering
	\vspace{-10pt}
	\centerline{\includegraphics[width=0.99\columnwidth]{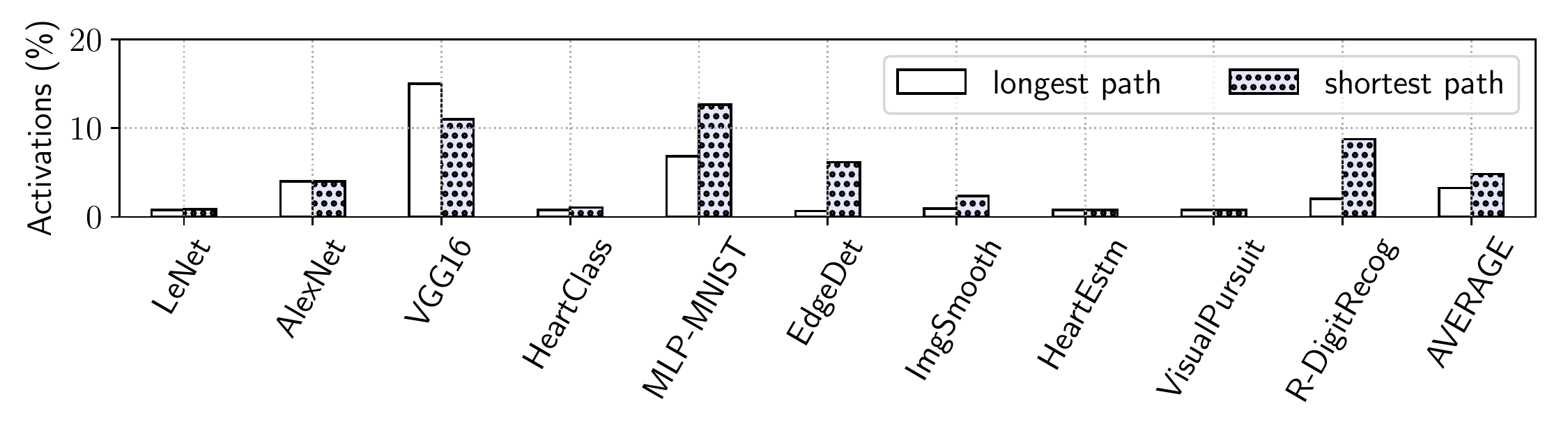}}
	\vspace{-10pt}
	\caption{Fraction of activation of memristor on the longest and shortest current paths in a crossbar using \prior{}~\cite{spinemap}.}
	\vspace{-10pt}
	\label{fig:activations}
\end{figure}

\mr{Figure~\ref{fig:activations} plots the number of synaptic activation, i.e., spikes propagating through the longest and the shortest current paths in a crossbar as fractions of the total synaptic activation.} Results are reported for 10 machine learning workloads (see Sec.~\ref{sec:evaluation}) using \prior{}~\cite{spinemap}.
We observe that the number of activation on the longest and shortest current paths are on average 3\% and 5\% of the total number of activation, respectively. 
Higher synaptic activation on shorter current paths in a crossbar can lead to lowering of the lifetime of memristors on those paths due to their lower cell endurance (see observations 1 and 2, and the endurance and lifetime formulations in Section~\ref{sec:endurance_model}). 


\section{Endurance Modeling}\label{sec:endurance_model}

\begin{figure}[h!]
	\centering
	\vspace{-10pt}
	\centerline{\includegraphics[width=0.99\columnwidth]{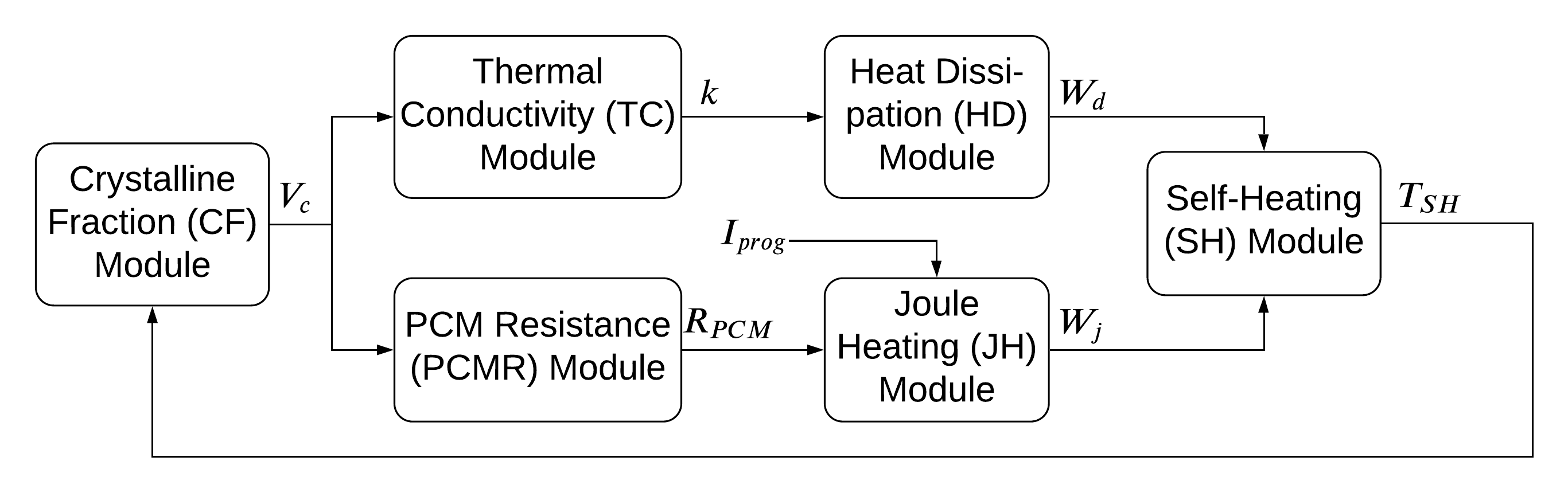}}
	\vspace{-10pt}
	\caption{Iterative approach to calculating the self-heating temperature of a PCM cell during amorphization.}
	\vspace{-10pt}
	\label{fig:tcal}
\end{figure}

We use the phenomenological endurance model~\cite{Strukov2016}, which computes endurance of a PCM cell as a function of its self-heating temperature obtained during amorphization of its crystalline state. Figure~\ref{fig:tcal} shows the iterative approach to compute this self-heating temperature (\ineq{T_{SH}})~\cite{xi2011spice,marcolini2013modeling}. 

At start of the amorphization process, the temperature of a PCM cell is equal to the ambient temperature \ineq{T_{amb}}. Subsequently, the PCM temperature is computed iteratively as follows. For a given crystalline fraction \ineq{V_C} of the GST material within the cell, the thermal conductivity \ineq{k} is computed using the \texttt{TC Module}, and PCM resistance \ineq{R_{PCM}} using the \texttt{PCMR Module}. The thermal conductivity is used to compute the heat dissipation \ineq{W_d} using the \texttt{HD Module}, while the PCM resistance is used to compute the Joule heating in the GST \ineq{W_j} for the programming current \ineq{I_{prog}} using the \texttt{JH Module}. The self-heating temperature \ineq{T_{SH}} is computed inside the \texttt{SH Module} using the Joule heating and the heat dissipation. Finally, the self-heating temperature is used to compute the crystallization fraction \ineq{V_c} using the \texttt{CF Module}. The iterative process terminates when the GST is amorphized, i.e., \ineq{V_c = 0}. We now describe these steps.

\begin{itemize}
    \item \textbf{Crystallization Fraction (CF) Module:} CF represents the fraction of solid in a GST during the application of a reset current. \ineq{V_c} is computed using the Johnson-Mehl-Avrami (JMA) equation
    as
    \begin{equation}
        \label{eq:vc}
        \footnotesize V_c=\text{exp}\left[-\alpha \times\frac{(T_{SH}-T_{amb})}{T_m}\times t\right],
    \end{equation}
    \mrr{where \ineq{t} is the time, \ineq{T_m = 810K}  is the melting temperature of the GST material~\cite{xi2011spice,marcolini2013modeling}, \ineq{T_{amb}} is the ambient temperature computed using~\cite{twisha_thermal,das2015reliability}, and \ineq{\alpha = 2.25} is a fitting constant~\cite{xi2011spice,marcolini2013modeling}.}
    \item \textbf{Thermal Conductivity (TC) Module:} TC of the GST is computed as~\cite{Liao2008}
    \begin{equation}
        \label{eq:k}
        \footnotesize k=(k_a-k_c)\times V_c+k_a,
    \end{equation}
    \mrr{where \ineq{k_a=0.002 WK^{-1}cm^{-1}} for amorphous GST, \ineq{k_c=0.005 WK^{-1}cm^{-1}} for crystalline GST~\cite{xi2011spice,marcolini2013modeling}.}
    \item \textbf{PCM Resistance (PCMR) Module:} The effective resistance of the PCM cell is given by
    \begin{equation}
        \label{eq:rpcm}
        \footnotesize R_{PCM}=R_{set}+(1-V_c)\times(R_{reset}-R_{set}),
    \end{equation}
    \mrr{where \ineq{R_{set} = 10K\Omega} in the crystalline state of the GST and \ineq{R_{reset} = 200K\Omega} in the amorphous state.}
    \item \textbf{Heat Dissipation (HD) Module:} Assuming heat is dispersed to the surrounding along the thickness of the PCM cell, HD is computed as~\cite{Kwong2008}
    \begin{equation}
        \label{eq:wd}
        \footnotesize W_d=\frac{kV}{l^2}(T_{SH}-T_{amb}),
    \end{equation}
    \mrr{where \ineq{l =120~nm} is the thickness and \ineq{V = 4\times 10^{-14} cm^3} is the volume of GST~\cite{xi2011spice,marcolini2013modeling}.}
    \item \textbf{Joule Heating (JH) Module:} The heat generation in a PCM cell due to the programming current \ineq{I_{prog}} is
    \begin{equation}
        \label{eq:wj}
        \footnotesize W_j=I_{prog}^2\times R_{PCM}.
    \end{equation}
    \item \textbf{Self-Heating (SH) Module:} The SH temperature of a PCM cell is computed by solving an ordinary differential equation as~\cite{xi2011spice}
    \begin{equation}
        \label{eq:tsh}
        \footnotesize T_{SH}=\frac{I_{prog}^2R_{PCM}l^2}{kV}-\left[1-\text{exp}\left(-\frac{kt}{l^2C}\right)\right]+T_{amb},
    \end{equation}
    \mrr{where \ineq{C = 1.25 JK^{-1} cm^{-3}} is the heat capacity of the GST~\cite{xi2011spice,marcolini2013modeling}.}
\end{itemize}

\mr{
The endurance of a PCM cell is computed as~\cite{Strukov2016}
\begin{equation}
    \label{eq:endurance_part_1}
    \footnotesize \text{Endurance}\approx \frac{t_f}{t_s},
\end{equation}
where \ineq{t_f} and \ineq{t_s} are respectively, the failure time and the switching time. In this model, to switch memory state of a PCM cell, an ion (electron) must travel a distance \ineq{d} across insulating matrix (the gate oxide) upon application of the programming current \ineq{I_{prog}}, which results in the write voltage \ineq{V} across the cell. Assuming thermally activated motion of an with activation energy \ineq{U_s} and local self-heating thermal temperature \ineq{T_{SH}}, the switching speed can be approximated as
\begin{equation}
    \label{eq:switching_time}
    \footnotesize t_s=\frac{d}{v_s}\approx \frac{2d}{fa}exp\left(\frac{U_s}{k_BT_{SH}}\right)exp\left(-\frac{qV}{2k_BT_{SH}}\frac{a}{d}\right),
\end{equation}
}
\mrr{where \ineq{d = 10 nm}, \ineq{a = 0.2 nm}, \ineq{f = 10^{13} Hz}, and \ineq{U_s = 2 eV}~\cite{Strukov2016}.}

\mr{
The failure time is computed considering that the endurance failure mechanism is due to thermally activated motion of ions (electrons) across the same distance \ineq{d} but with higher activation energy \ineq{U_F}, so that the average time to failure is
\begin{equation}
    \label{eq:failure_time}
    \footnotesize t_f=\frac{d}{v_f}\approx \frac{2d}{fa}exp\left(\frac{U_f}{k_BT_{SH}}\right)exp\left(-\frac{qV}{2k_BT_{SH}}\frac{a}{d}\right) 
\end{equation}
}
\mrr{where \ineq{U_f = 3ev}~\cite{Strukov2016}.}

\mr{
The endurance, which is the ratio of average failure time and switching time, is given by
\begin{equation}
    \label{eq:endurance}
    \footnotesize \text{Endurance}\approx \frac{t_f}{t_s}\approx \text{exp}\left(\frac{\gamma}{T_{SH}}\right),
\end{equation}
}
\mrr{where \ineq{\gamma = 1000} is  a fitting parameter~\cite{Strukov2016}.}

\mrr{The thermal and endurance models are used in our SNN mapping framework to improve endurance of neuromorphic hardware platforms (see Section~\ref{sec:results}). Although we have demonstrated our proposed SNN mapping approach using these models (see Section~\ref{sec:mapping}), the mapping approach can be trivially extended to incorporate other published models.}

\subsection{Model Prediction}
\mrr{
The thermal and endurance models in Equations~\ref{eq:tsh} and \ref{eq:endurance}, respectively are integrated as follows. The self-heating temperature of Equation~\ref{eq:tsh} is first computed using the PCM's programming current. This self-heating temperature is then used to compute the endurance using Equation~\ref{eq:endurance}.
}

\begin{figure}[h!]%
    \centering
    \vspace{-10pt}
    \subfloat[Change in crystallization fraction in PCM\label{fig:cf}.]{{\includegraphics[width=9cm]{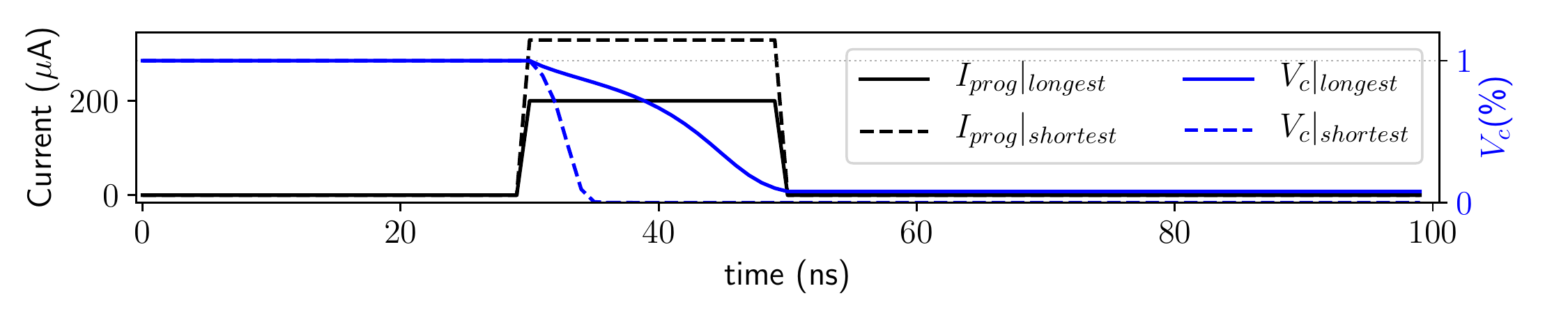} }}%
    \qquad
    \subfloat[Change in PCM resistance\label{fig:pcmr}.]{{\includegraphics[width=9cm]{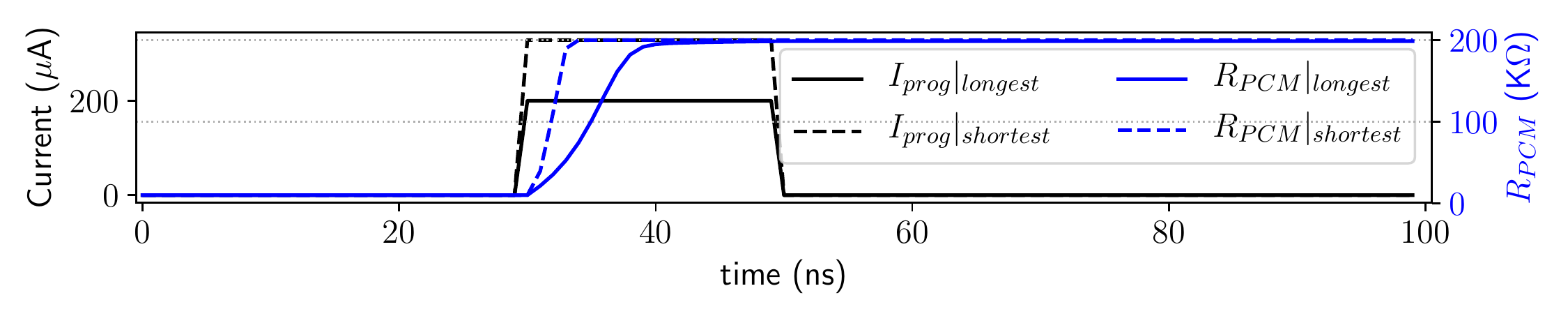} }}%
    \qquad
    \subfloat[Change in PCM temperature\label{fig:pcmt}.]{{\includegraphics[width=9cm]{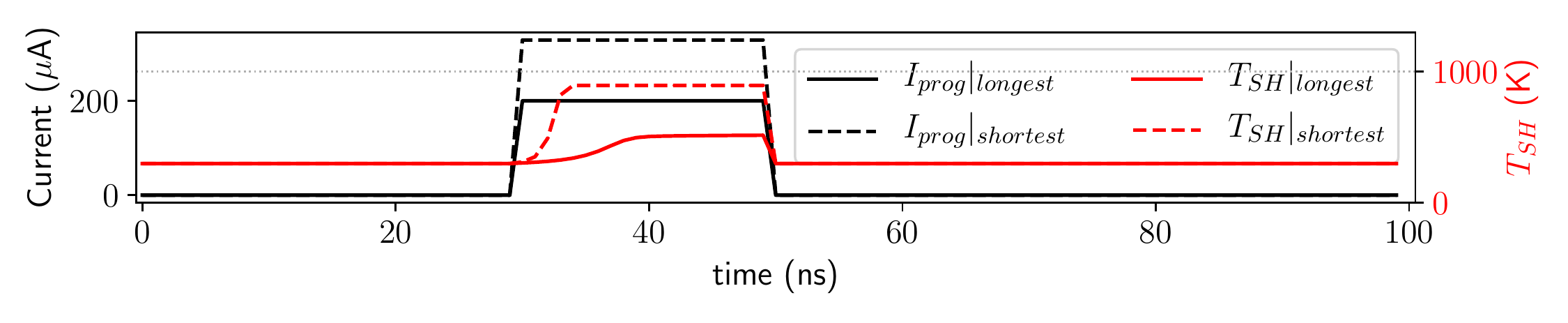} }}%
    \caption{Validation of the proposed model.}%
    \label{fig:model_validation}%
\end{figure}

\mr{Figure~\ref{fig:model_validation} shows the simulation of the proposed model with programming currents of \ineq{200\mu A} and \ineq{329 \mu A}, which correspond to the longest and shortest current paths in a 65nm 128x128 PCM crossbar at 298K.} Figures~\ref{fig:cf}, \ref{fig:pcmr}, and \ref{fig:pcmt} plot respectively, the crystallization fraction, the PCM resistance, and the temperature for these two current values. We make the following two key observations.

First, the speed of amorphization depends on the current, i.e., with higher programming current, the GST material amorphizes faster. This means that the PCM cells on shorter current paths are faster to program.
Second, the self-heating temperature is higher for higher programming current. This means that PCM cells on shorter current paths have lower endurance. 

\begin{figure}[h!]%
    \centering
    \subfloat[Thermal map for PCM RESET operations in a 128x128 crossbar\label{fig:thermal_map}.]{{\includegraphics[width=4.15cm]{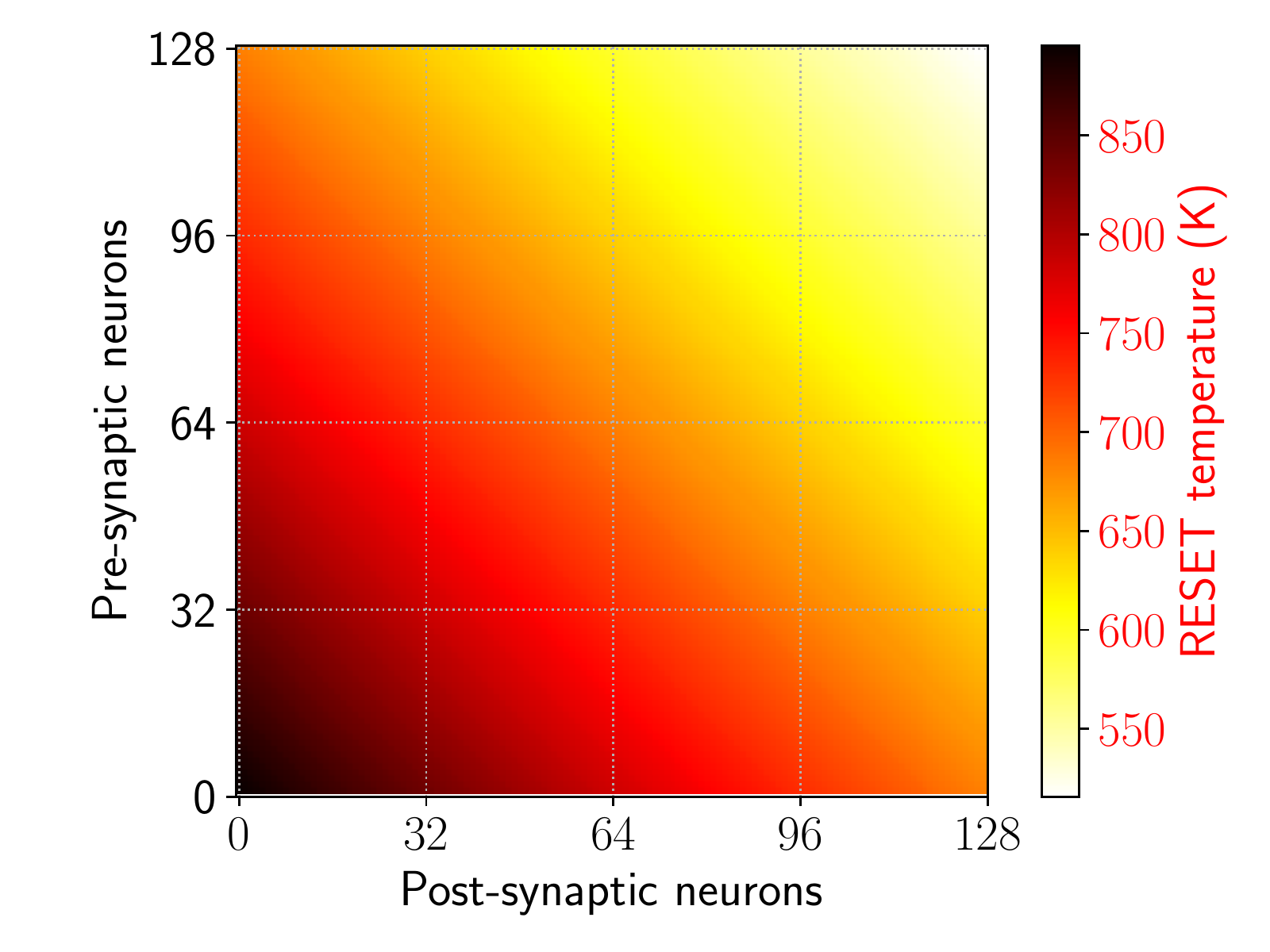} }}%
    \quad
    \subfloat[Endurance map of the PCM cells in a 128x128 crossbar\label{fig:endurance_map}.]{{\includegraphics[width=4.15cm]{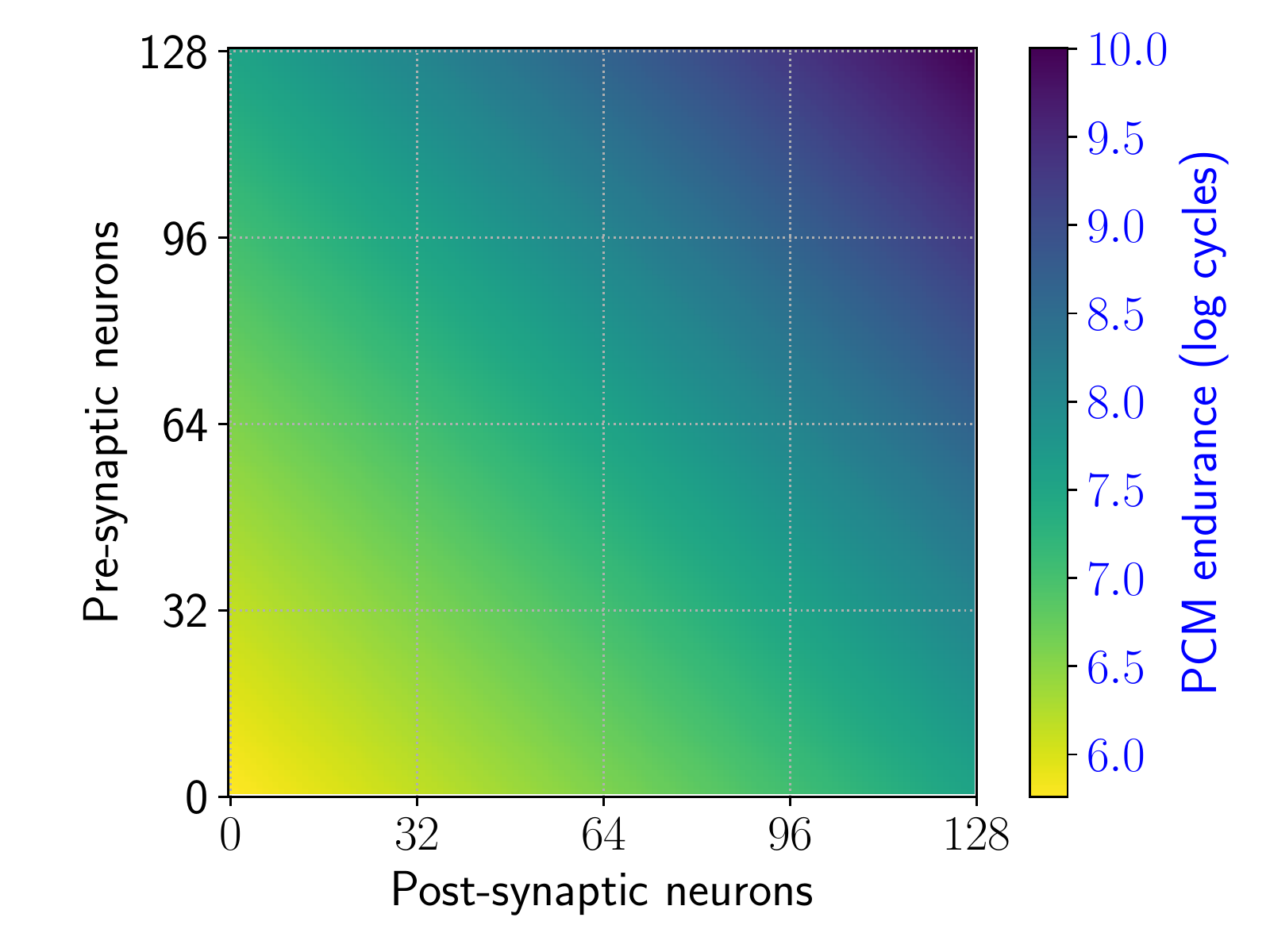} }}%
    \caption{Temperature and endurance map of a 128x128 crossbar at \ineq{65}nm process node with \ineq{T_{amb} = 298K}.}%
    \label{fig:temperature_endurance_map}%
\end{figure}

\mrr{Figure~\ref{fig:model_validation} is consistent with the change in crystallization volume, resistance, and self-heating temperature in PCM cells as reported in~\cite{xi2011spice,marcolini2013modeling}.}
Figure~\ref{fig:temperature_endurance_map} plots the temperature and endurance maps of a 128x128 crossbar at {65}nm process node with \ineq{T_{amb} = 298K}. The PCM cells at the bottom-left corner have higher self-heating temperature than at the top-right corner. This asymmetry in the self-heating temperature creates a wide distribution of endurance, ranging from \ineq{10^6} cycles for PCM cells at the bottom-left corner to \ineq{10^{10}} cycles at the top-right corner. \mrr{These endurance values are consistent with the values reported for recent PCM chips from IBM~\cite{burr2016recent}.}

Our goal is to assign synapses with higher activation towards the top-right corner using an intelligent SNN mapping technique, which we describe next.

\section{Endurance-Aware Intelligent Mapping}\label{sec:mapping}
We present \tech{}, our novel endurance-aware technique to map SNNs to neuromorphic hardware. To this end, we first formulate a joint metric \emph{effective lifetime} (\ineq{\mathcal{L}_{i,j}}), defined for the memristor connecting the \ineq{i^\text{th}} pre-synaptic neuron with \ineq{j^\text{th}} post-synaptic neuron in a memristive crossbar as
\begin{equation}
    \label{eq:normalized_lifetime}
    \footnotesize \mathcal{L}_{i,j} =  \mathcal{E}_{i,j}/a_{i,j}, 
\end{equation}
where \ineq{a_{i,j}} is the number of synaptic activations of the memristor in a given SNN workload and \ineq{\mathcal{E}_{i,j}} is its endurance. Equation~\ref{eq:normalized_lifetime} combines the effect of software (SNN mapping) on hardware (endurance and temperature). \tech{} aims to maximize the minimum normalized lifetime, i.e.,
\begin{equation}
    \label{eq:optimization_objective}
    \footnotesize F_\text{opt} = \text{maximize} \{\min_{i,j} \mathcal{L}_{i,j}\}
\end{equation}

In most earlier works on wear-leveling in the context of non-volatile main memory (e.g., Flash), lifetime is computed in terms of utilization of NVM cells, ignoring the variability of endurance within the device. Instead, we formulate the effective lifetime by considering a memristor's endurance and its utilization in a workload. This is to allow cells with higher endurance to have higher utilization in a workload.

\subsection{High-level Overview}
\begin{figure}[h!]
	\centering
	\vspace{-10pt}
	\centerline{\includegraphics[width=0.99\columnwidth]{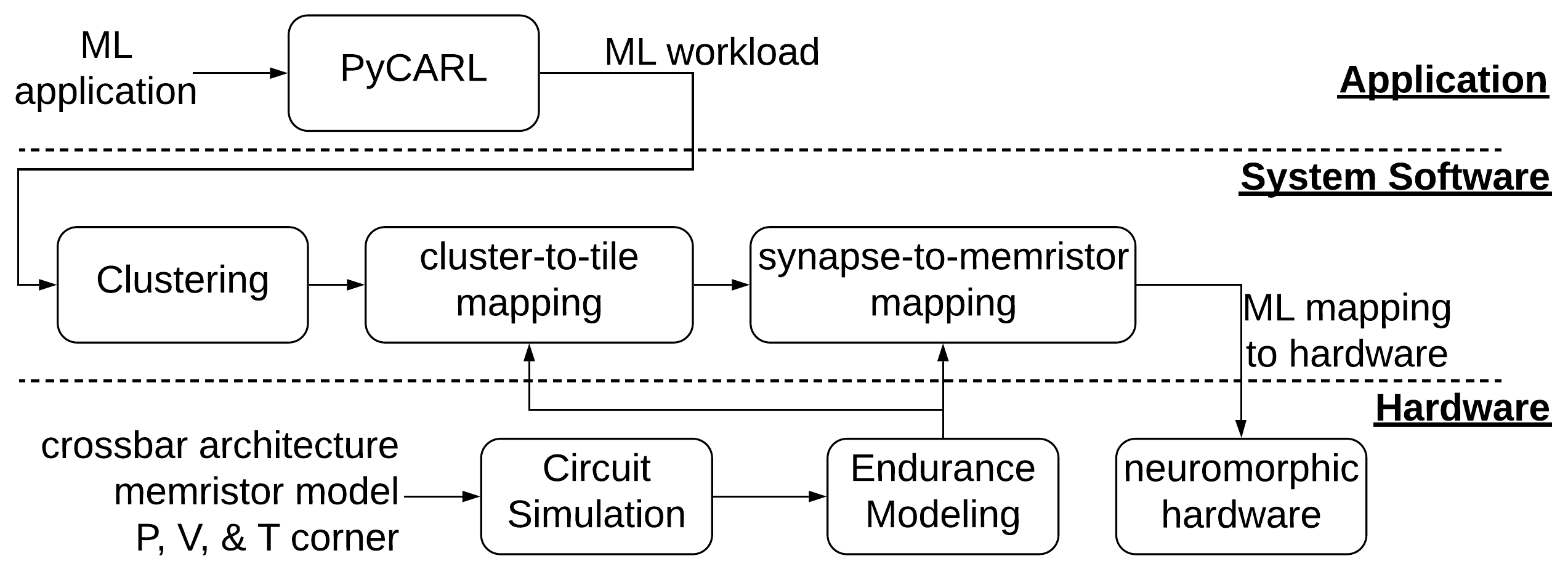}}
	\vspace{-10pt}
	\caption{High-level overview of \tech{}.}
	\vspace{-10pt}
	\label{fig:overview}
\end{figure}

Figure~\ref{fig:overview} shows a high-level overview of \tech{}, consisting of three abstraction layers -- the application layer, system software layer, and hardware layer. A machine learning application is first simulated using PyCARL~\cite{pycarl}, which uses CARLsim~\cite{carlsim} for training and testing of SNNs. PyCARL estimates spike times and synaptic strength on every connection in an SNN. This constitutes the workload of the machine learning application. 
\tech{} maps and places neurons and synapses of a workload to crossbars of a neuromorphic hardware, improving the effective lifetime. To this end, a machine learning workload is first analyzed to generate clusters of neurons and synapses, where each cluster can fit on a crossbar. \fix{\tech{} uses the Kernighan-Lin Graph Partitioning algorithm of \prior{}~\cite{spinemap} to partition an SNN workload, minimizing the inter-cluster spike communication (see Table~\ref{tab:spinemap_vs_esping}  for comparison of \tech{} with \prior{}). By reducing the inter-cluster communication, \tech{} reduces the energy consumption and latency on the shared interconnect (see Sec.~\ref{sec:energy_results}).} Next, \tech{} uses an instance of the Particle Swarm Optimization (PSO)~\cite{kennedy2010particle} to map the clusters to the tiles of a hardware, maximizing the minimum effective lifetime of memristors (Equation~\ref{eq:normalized_lifetime}) in each tile's crossbar. Synapses of a cluster are implemented on memristors using the synapse-to-memristor mapping, ensuring that those with higher activation are mapped to memristors with higher endurance, and vice versa. 

To perform the optimization using PSO, \tech{} uses crossbar specification, including its dimensions, architecture, and memristor technology, and performs circuit simulations at a target P, V, and T corner. Extracted currents in the crossbar are used in the endurance model (see Sec.~\ref{sec:endurance_model}) to generate the endurance map, which is then used in the cluster-to-tile and synapse-to-memristor mapping, optimizing the effective lifetime.

\fix{
Table~\ref{tab:spinemap_vs_esping} reports the differences between the objective function of \prior{} and \tech{}. In addition to the comparison between \prior{} and \tech{}, we also show the performance of a hybrid approach \priorpp{} (see Fig.~\ref{fig:sota}), which uses the synapse-to-memristor mapping of \tech{} with \prior{}. See our results in Section~\ref{sec:results}.
}
\begin{table}[h!]
	\renewcommand{\arraystretch}{1.0}
	\setlength{\tabcolsep}{1.2pt}
	\caption{\tech{} vs. \prior{}~\cite{spinemap}.}
	\label{tab:spinemap_vs_esping}
	\vspace{-10pt}
	\centering
	{\fontsize{6}{9}\selectfont
		\begin{tabular}{rc|>{\centering\arraybackslash}m{2.5cm}|>{\centering\arraybackslash}m{2.5cm}}
			\hline
			&& \textbf{\prior{}~\cite{spinemap}} & \textbf{\tech{} (proposed)}\\
			\hline
			\multirow{2}{*}{Clustering} & Algorithm & Kernighan-Lin Graph Partitioning~\cite{kernighan1970efficient} & Kernighan-Lin Graph Partitioning~\cite{kernighan1970efficient}\\
			\cline{2-4}
			& Objective & Energy & Energy\\
            \hline
             \multirow{2}{*}{Cluster-to-Tile} & Algorithm & PSO & PSO \\
             \cline{2-4}
             & Objective & Energy & Effective Lifetime\\
             \hline
             \multirow{2}{*}{Synapse-to-Memristor} & Algorithm & \multirow{2}{*}{---} & Sorting heuristic\\
             \cline{2-2}
             \cline{4-4}
             & Objective & & Effective Lifetime \\
             \hline
	    \end{tabular}
	 }
\end{table}

\fix{
Although PSO is previously proposed in \prior{}, our novelty is in the use of the proposed synapse-to-memristor mapping step, which is integrated inside each PSO iteration to find the minimum effective lifetime.
}

\subsection{Heuristic-based Synapse-to-Memristor Mapping}
\begin{figure}[h!]
	\centering
	\vspace{-10pt}
	\centerline{\includegraphics[width=0.99\columnwidth]{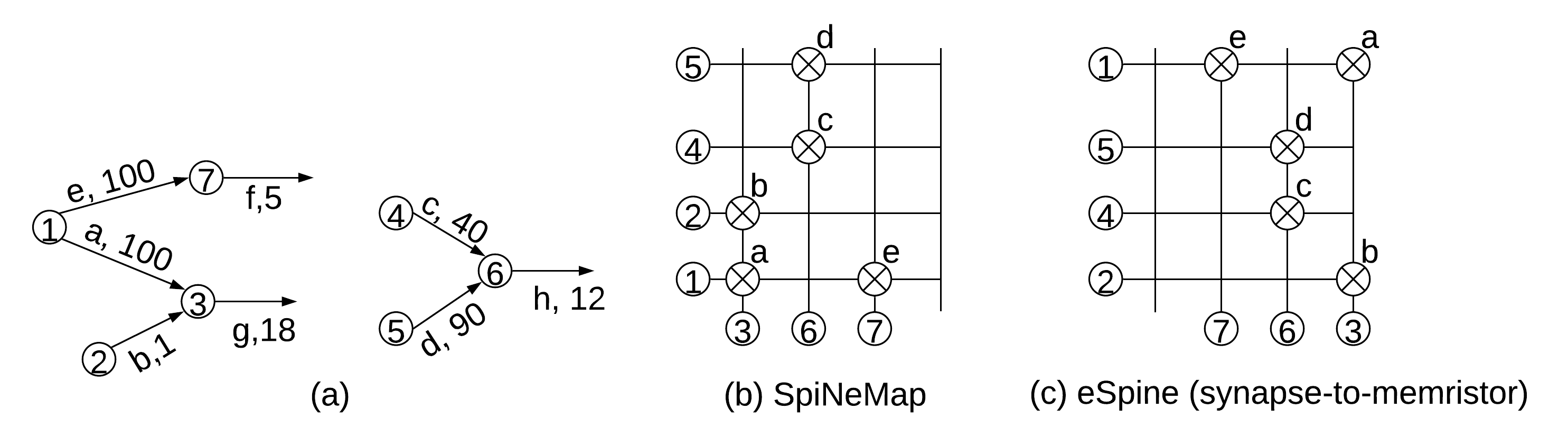}}
	\vspace{-10pt}
	\caption{Synapse-to-memristor mapping of \tech{}.}
	\vspace{-10pt}
	\label{fig:synapse-to-memristor}
\end{figure}

Figure~\ref{fig:synapse-to-memristor} illustrates the synapse-to-memristor mapping of \tech{} and how it differs from \prior{}.
\mr{
Figure~\ref{fig:synapse-to-memristor}a illustrates the implementation of four pre-synaptic and three post-synaptic neurons on a 4x4 crossbar.
}
The letter and number on a connection indicate the synaptic weight and number of activation, respectively. Existing technique such as \prior{} maps synapses arbitrarily on memristors. As a result, a synapse with higher activation may get placed at the bottom-left corner of a crossbar where memristors have lower endurance (see Fig.~\ref{fig:synapse-to-memristor}b). 
\tech{}, on the other hand, incorporates the endurance variability in its synapse-to-memristor mapping process. It first sorts pre-synaptic neurons based on their activation, and then allocates them such that those with higher activation are placed at the top-right corners, where memristors have higher endurance (see Fig.~\ref{fig:synapse-to-memristor}c).
\mr{
Once the pre-synaptic neurons are placed along the rows, the post-synpatic neurons are placed along the columns, considering their connection to the pre-synaptic neurons, and their activation. In other words, post-synaptic neurons with higher activation are placed towards the right corner of a crossbar. This is shown in Fig.~\ref{fig:synapse-to-memristor}c, where the post-synaptic neuron 7 (with 5 activation) is mapped to the left of the post-synaptic neuron 3 (with 18 activation), both of which receives input from the same pre-synaptic neuron 1. 
This is done to incorporate the online weight update mechanism in SNNs, which depend on both the pre- and post-synaptic activation (see Section~\ref{sec:scope}).
}
This synapse-to-memristor mapping is part of Alg.~\ref{alg:max_lifetime} (lines 9-10).

\subsection{PSO-based Cluster-to-Tile Mapping}
To formulate the PSO-based optimization problem, let \ineq{{G} ({C,S})} be a machine learning workload with a set \ineq{{C}} of clusters and a set \ineq{{S}} of connections between the clusters. The workload is to be executed on a hardware \ineq{H(T,L)} with a set \ineq{T} of tiles (each tile has one crossbar) and a set \ineq{L} of links between the tiles. Mapping of the application \ineq{G} to the hardware \ineq{H}, \ineq{\mathcal{M} = \{m_{x,y}\}} is defined as
\begin{equation}
    \label{eq:mapping_rep}
    \footnotesize m_{x,y} = \begin{cases}
    1 & \text{if cluster } {c}_x \in {C} \text{ is mapped to tile } {t}_y\in{T}\\
    0 & \text{otherwise}
    \end{cases}
\end{equation}

Algorithm~\ref{alg:max_lifetime} computes the minimum effective lifetime of all memristors in the hardware for a given mapping \ineq{\mathcal{M}}. 

\begin{algorithm}[h]
	\scriptsize{
	    \KwIn{$\mathcal{M}$}
	    \KwOut{$\mathcal{L}$}
	    \For{$t_y\in{T}$ \tcc{iterate for each tile in the hardware}}{
			$ S_y = \{c_x\}\ni m_{x,y} = 1$\tcc{clusters mapped to \ineq{t_y}}
			$\mathcal{L}_{i,j}^y = 0~\forall~\{i,j\}\in 1,2,\cdots,M$\tcc{Initialize the effective lifetime on tile $t_y$.}
			\For{$c_k\in{S_y}$ \tcc{iterate for each cluster}}{
			    $N_k = \{n\}$\tcc{pre-synaptic neurons of \ineq{c_k}}
			    $A_k = \{a\}$\tcc{number of activations of \ineq{n}}
			    sort $A_k$\tcc{sort the pre-synaptic neurons in descending order of their activations.}
			    map $N_k$ to the crossbar using sorted $A_k$\tcc{place the pre-synaptic neurons sorted by their activations starting from the farthest input in the crossbar.}
			    repeat lines 7-10 for post-synaptic neurons\;
			    $\mathcal{L}_{i,j}^y = \mathcal{L}_{i,j}^y + \mathcal{E}_{i,j}/a_{i,j}$ \tcc{using Equation~\ref{eq:normalized_lifetime}}
			}
			$\mathcal{L}_y = \texttt{min}\{\mathcal{L}_{i,j}^y\}$\tcc{minimum effective lifetime}
		}
		\textbf{return} \texttt{min}$\{\mathcal{L}_y\}$\tcc{return minimum effective lifetime of all crossbars}
	}
	\caption{\small \texttt{MinEffLife}(): Compute minimum effective lifetime of crossbars for mapping \ineq{\mathcal{M}}.}
	\label{alg:max_lifetime}
\end{algorithm}

For each tile, the algorithm first records all clusters mapped to the tile in the set \ineq{S_y} (line 3), and initializes the effective lifetime of the crossbar on the tile (line 4). For each cluster mapped to the tile, the algorithm records all its pre-synaptic neurons in the set \ineq{N_k} (line 7) and their activation, i.e., the number of spikes in the set \ineq{A_k} (line 8). The two sets are sorted in descending order of \ineq{A_k} (line 9). Next, the cluster (i.e, pre-synaptic neurons, post-synaptic neurons, and their synaptic connections) is placed on the crossbar (line 10-11). To do so, pre-synaptic neurons with higher activation are mapped farther from the origin (see Fig.~\ref{fig:synapse-to-memristor}) to ensure they are on longer current paths. This is to incorporate the endurance variability within each crossbar. \mr{The post-synaptic neurons are mapped along the columns by sorting their activation.} With this mapping, the effective lifetime is computed (line 12). The minimum effective lifetime is retained (line 14). The algorithm is repeated for all tiles of the hardware. Finally, the minimum effective lifetime of all crossbars in the hardware is returned (line 16).

The \textbf{fitness function} of \tech{} is
\begin{equation}
    \label{eq:fitness_function}
    \footnotesize F = \texttt{MinEffLife} (\mathcal{M})
\end{equation}

The \textbf{optimization objective} of \tech{} is
\begin{equation}
    \label{eq:opt_obj}
    \footnotesize \mathcal{L}_\text{min} = \mathcal{L}_\text{a}, \text{ where } a = \argmin\{\text{MinEffLife}(\mathcal{M}_i) | i\in 1,2,\cdots\},
\end{equation}

The constraint to this optimization problem is that a cluster can map to exactly 1 tile, i.e.,
    \begin{equation}
        \label{eq:const_1}
        \footnotesize \sum_y m_{x,y} = 1~\forall~x
    \end{equation}

To solve Equation~\ref{eq:opt_obj} using PSO, we instantiate \ineq{n_p} swarm particles. The position of these particles are solutions to the fitness functions, and they represent cluster mappings, i.e., \ineq{\mathcal{M}}'s in Equation \ref{eq:opt_obj}. Each particle also has a velocity with which it moves in the search space to find the optimum solution. During the movement, a particle updates its position and velocity according to its own experience (closeness to the optimum) and also experience of its neighbors. We introduce the following notations.

\vspace{-10pt}
\begin{footnotesize}
 	\begin{align}
 	\label{eq:pso_defn}
 	D = |\mathcal{C}|\times|\mathcal{V}| &= \text{dimensions of the search space}\\
 	\mathbf{\Theta} = \{\mathbf{\theta}_l\in\mathbb{R}^{D}\}_{l=0}^{n_p-1} &= \text{positions of particles in the swarm}\nonumber \\
 	\mathbf{V} = \{\mathbf{v}_l\in\mathbb{R}^{D}\}_{l=0}^{n_p-1} &= \text{velocity of particles in the swarm}\nonumber 
 	\end{align}
 \end{footnotesize}\normalsize
Position and velocity of swarm particles are updated, and the fitness function is computed as

\vspace{-10pt}
\begin{footnotesize}
	\begin{align}
	\label{eq:pos_vel_update}
	\mathbf{\Theta}(t+1) &= \mathbf{\Theta}(t) + \mathbf{V}(t+1)\\
	\mathbf{V}(t+1) &= \mathbf{V}(t) + \varphi_1\cdot\Big(P_{\text{best}}-\mathbf{\Theta}(t)\Big) + \varphi_2\cdot\Big(G_{\text{best}}-\mathbf{\Theta}(t)\Big)\nonumber\\
	F(\theta_l) &= \mathcal{L}_l = \text{MinEffLife}(M_l)\nonumber
	\end{align}
\end{footnotesize}
\normalsize where $t$ is the iteration number, $\varphi_1,\varphi_2$ are constants and $P_{\text{best}}$ (and $G_{\text{best}}$) is the particle's own (and neighbors) experience. 
Finally, local and global bests are updated as

\vspace{-10pt}
\begin{footnotesize}
    \begin{align}
    \label{eq:pbest}
        && P_\text{best}^l = F(\theta_l) \text{ if } F(\theta_l) < F(P_\text{best}^l)\nonumber \\
        && G_\text{best} = \displaystyle \argmin_{l=0,\dots n_p-1} P_\text{best}^l
    \end{align}
\end{footnotesize}\normalsize

Due to the binary formulation of the mapping problem (see Equation \ref{eq:mapping_rep}), we need to binarize the velocity and position of Equation \ref{eq:pso_defn}, which we illustrate below.

\vspace{-10pt}
\begin{footnotesize}
	\begin{align}
	\label{eq:binarization}
	&\hat{\mathbf{V}} = \texttt{sigmoid}(\mathbf{V}) = \frac{1}{1+\texttt{e}^{-\mathbf{V}}} \nonumber \\
	&	\hat{\Theta} = \begin{cases}
	0 \text{~~if } \texttt{rand()} < \hat{\mathbf{V}}\\
	1 \text{~~otherwise }
	\end{cases}
	\end{align}
\end{footnotesize}

\begin{figure}[h!]
	\centering
	\vspace{-10pt}
	\centerline{\includegraphics[width=0.66\columnwidth]{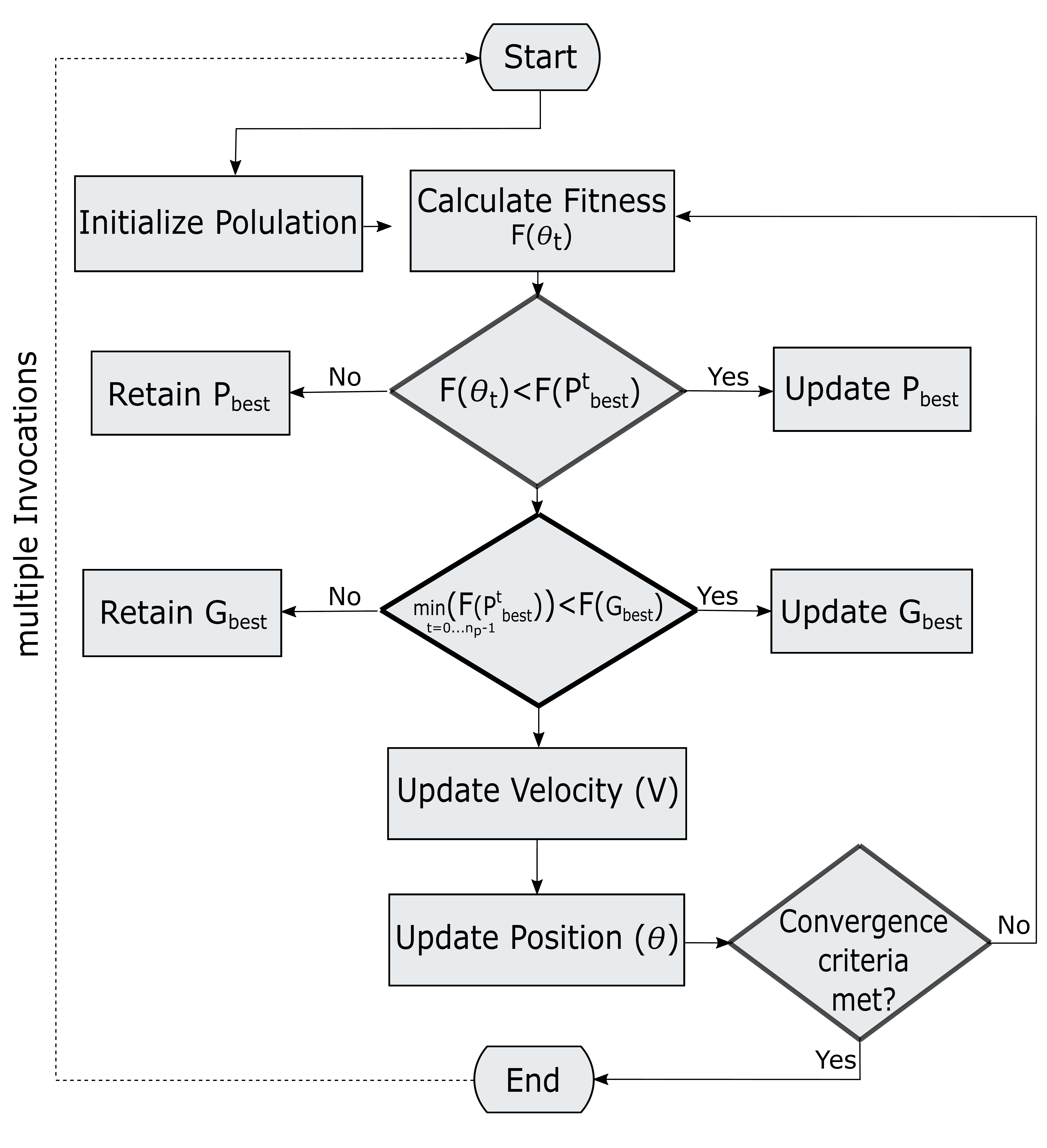}}
	\vspace{-10pt}
	\caption{Flow chart of our PSO algorithm.}
	\vspace{-10pt}
	\label{fig:pso_flow}
\end{figure}

Figure \ref{fig:pso_flow} illustrates the PSO algorithm. The algorithm first initializes positions of the PSO particles (\ref{eq:mapping_rep}). Next, the algorithm runs for \ineq{N_\text{PSO}} iterations. At each iteration, the PSO algorithm evaluates the fitness function (\ineq{F}) and updates its position based on the local and global best positions (Equation \ref{eq:pos_vel_update}), binarizing these updates using Equation \ref{eq:binarization}.

\mr{The PSO algorithm of \tech{} can be used to explore the energy and lifetime landscape of different neuron mapping solutions to the hardware.
Section~\ref{sec:energy_tradeoffs} illustrates such exploration for a representative application. 
\tech{} gives designers the flexibility to combine energy and lifetime metrics beyond simply obtaining the minimum energy and maximum lifetime mappings (for instance, minimizing energy for a given lifetime target, and vice versa).
}



\section{\mr{Extended Scope of \tech{}}}\label{sec:extended_scope}
\subsection{\mr{Other Memristor Technologies}}\label{sec:other_memristors}
\mr{
Temperature-related endurance issues are also critical for other memristor technologies such as FeRAM and STT-/SOT-MRAM. A thermal model for Magnetic Tunnel Junction (MTJ), the basic storage element in STT-MRAM based memoristor, is proposed in~\cite{zhang2018addressing}. According to this model, the self-heating temperature is due to the spin polarization percentages of the free layer and the pinned layer in the MTJ structure, which are dependent on the programming current. Similarly, a thermal model for FeRAM-based memristor is proposed in~\cite{gupta2020temperature}. These models can be incorporated directly into our SPICE-level crossbar model to generate the thermal and endurance maps, similar to those presented in Figure~\ref{fig:temperature_endurance_map} for PCM. The proposed cluster-to-tile mapping and the synapse-to-crossbar mapping (see Section~\ref{sec:mapping}) can then use these maps to optimize the placement of synapses for a target memristor technology, improving its endurance.
Although the exact numerical benefit may differ, \tech{} can improve endurance for different memristor technologies. 
}

\subsection{\mr{Other Reliability Issues}}
\mrr{
There are other thermal-related reliability issues in memristors, for instance retention-time~\cite{stanisavljevic2015phase,bhattacharjee2020rethinking,zyarah2018semi,santos2014criticality} and transistor circuit aging~\cite{zhang2019aging,reneu,frameworkCAL,NeuromorphicLR,das2018reliable,das2012fault,das2015workload,das2013communication,das2014communication}. Retention time is defined as the time for which a memristor can retain its programmed state. Recent studies show that retention time reduces significantly with increase in temperature~\cite{stanisavljevic2015phase}.
}
\mr{
Retention time issues are relevant for supervised machine learning, where the synaptic weights are programmed on memristors once, during inference. For online learning (which is the focus of this work), synaptic weight update frequency is usually much smaller than the retention time. Therefore, a reduction in retention time is less of a concern. Nevertheless, by lowering the average temperature of crossbars, \tech{} also addresses the retention time-related reliability concerns.
}

\section{Evaluation Methodology}\label{sec:evaluation}
\subsection{\mr{Use-Case of \tech{}}}\label{sec:scope}
\begin{figure}[h!]
	\centering
	\vspace{-10pt}
	\centerline{\includegraphics[width=0.99\columnwidth]{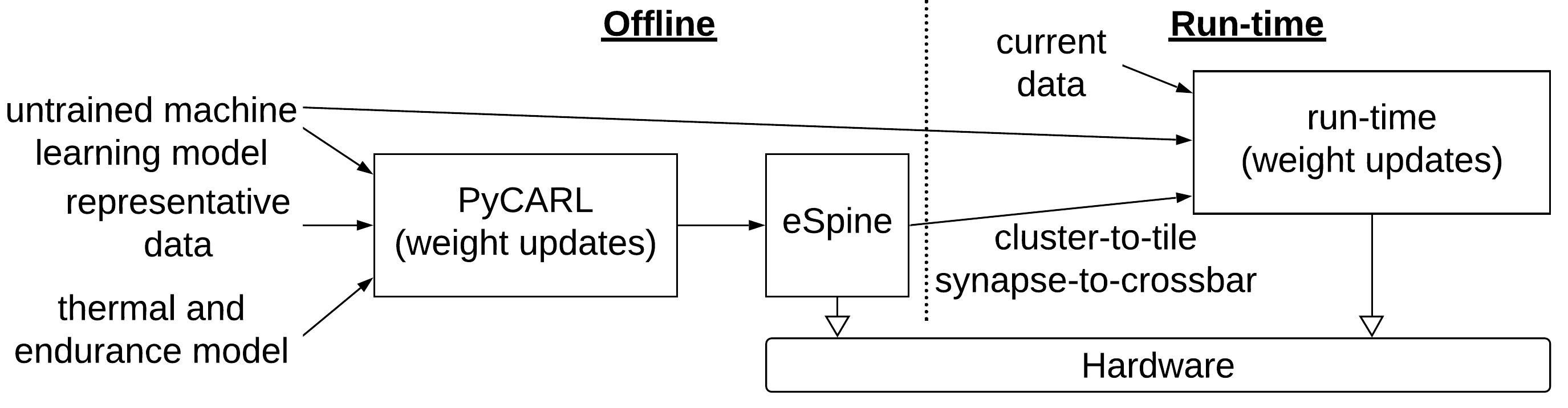}}
	\vspace{-10pt}
	\caption{Use-Case of \tech{}.}
	\vspace{-10pt}
	\label{fig:scope}
\end{figure}

\mr{
Figure~\ref{fig:scope} illustrates the use-case of \tech{} applied for on-line machine learning. We 
use Spike-Timing Dependent Plasticity (STDP)~\cite{dan2004spike}, which is an unsupervised learning algorithm for SNNs, where the synaptic weight between a pre- and a post-synaptic neuron is updated based on the timing of pre-synaptic spikes relative to the post-synaptic spikes.\footnote{Apart from STDP, many other online learning algorithms depend on the activation of both the pre- and post-synaptic neurons.}
STDP is typically used in online settings to improve accuracy of machine learning tasks. 
}

\mr{
A machine learning model is first analyzed offline using PyCARL with representative workload and data set. This is to estimate the relative activation frequency of the neurons in the model when it is trained at run-time using current data.
Although neuron activation can deviate at run-time, 
our more detailed analysis shows that using representative workload and data set, such deviations can be limited to only a few neurons in the model.\footnote{\mr{In the worst-case, the lifetime obtained using \tech{} for these few neurons will be similar to \prior{}. However, for most neurons in the model, \tech{} significantly outperforms \prior{}. Therefore, the lifetime obtained using \tech{} is higher (see Section~\ref{sec:lifetime_results}).}}
We have validated this observation for the evaluated applications that use ECG and image data (see Section~\ref{sec:evaluation}).
}

\mr{
The activation information obtained offline is processed using \tech{} (see Figure~\ref{fig:overview} for the details of \tech{}) to generate cluster-to-tile and synapse-to-crossbar mappings. 
The offline trained weight updates are discarded to facilitate relearning of the model from current (in-field) data. The untrained machine learning model is placed onto the hardware using the mappings generated from \tech{}.
}

\mr{
Although online learning is the main focus, \tech{} is also relevant for supervised machine learning, where no weight updates happen at run-time.
By mapping the most active neurons to the farthest corner of a crossbar (i.e., on longest current paths), \tech{} minimizes crossbar temperature, which reduces 1) leakage current and 2) circuit aging.
}

\subsection{Evaluated Applications}
We evaluate 10 SNN-based machine learning applications that are representative of three most commonly-used neural network classes --- convolutional neural network (CNN), multi-layer perceptron (MLP), and recurrent neural network (RNN).
These applications are 
1) LeNet based handwritten digit recognition with \ineq{28 \times 28} images of handwritten digits from the MNIST dataset;
2) AlexNet for ImageNet classification;
3) VGG16, also for ImageNet classification;
4) ECG-based heart-beat classification (HeartClass)~\cite{HeartClassJolpe,das2018heartbeat} using electrocardiogram (ECG) data;
5) {multi-layer perceptron (MLP)-based handwritten digit recognition} (MLP-MNIST)~\cite{Diehl2015} using the MNIST database;
6) {edge detection} (EdgeDet)~\cite{carlsim} on $64 \times 64$ images using difference-of-Gaussian; 
7) {image smoothing} (ImgSmooth)~\cite{carlsim} on $64 \times 64$ images; 
8) {heart-rate estimation} (HeartEstm)~\cite{HeartEstmNN} using ECG data;
9) RNN-based predictive visual pursuit (VisualPursuit)~\cite{Kashyap2018}; and
10) recurrent digit recognition (R-DigitRecog)~\cite{Diehl2015}.
Table~\ref{tab:apps} summarizes the topology, the number of neurons and synapses of these applications, and their baseline accuracy on DYNAP-SE using SpiNeMap~\cite{spinemap}.

\vspace{-10pt}
\begin{table}[h!]
	\renewcommand{\arraystretch}{0.8}
	\setlength{\tabcolsep}{2pt}
	\caption{Applications used to evaluate \tech{}.}
	\label{tab:apps}
	\vspace{-10pt}
	\centering
	\begin{threeparttable}
	{\fontsize{6}{10}\selectfont
		\begin{tabular}{cc|ccl|c}
			\hline
			\textbf{Class} & \textbf{Applications} & \textbf{Synapses} & \textbf{Neurons} & \textbf{Topology} & \textbf{Accuracy}\\
			\hline
			\multirow{4}{*}{CNN} & LeNet & 282,936 & 20,602 & CNN & 85.1\%\\
			& AlexNet & 38,730,222 & 230,443 & CNN & 90.7\%\\
			& VGG16 & 99,080,704 & 554,059 & CNN & 69.8 \%\\
			& HeartClass~\cite{HeartClassJolpe} & 1,049,249 & 153,730 & CNN & 63.7\%\\
			\hline
			\multirow{3}{*}{MLP} & DigitRecogMLP & 79,400 & 884 & FeedForward (784, 100, 10) & 91.6\%\\
			& EdgeDet \cite{carlsim} & 114,057 &  6,120 & FeedForward (4096, 1024, 1024, 1024) & 100\%\\
			& ImgSmooth \cite{carlsim} & 9,025 & 4,096 & FeedForward (4096, 1024) & 100\%\\
			\hline
 			\multirow{3}{*}{RNN} & HeartEstm \cite{HeartEstmNN} & 66,406 & 166 & Recurrent Reservoir & 100\%\\
 			& VisualPursuit \cite{Kashyap2018} & 163,880 & 205 & Recurrent Reservoir & 47.3\%\\
 			& R-DigitRecog \cite{Diehl2015} & 11,442 & 567 & Recurrent Reservoir & 83.6\%\\
			\hline
	\end{tabular}}
	\end{threeparttable}
	\vspace{-10pt}
\end{table}

\subsection{Hardware Models}
We model the DYNAP-SE neuromorphic hardware~\cite{dynapse} with the following configurations.

\begin{itemize}
    \item A tiled array of 4 tiles, each with a 128x128 crossbar. There are 65,536 memristors per crossbar.
    \item Spikes are digitized and communicated between cores through a mesh routing network using the Address Event Representation (AER) protocol.
    \item Each synaptic element is a PCM-based memristor. 
\end{itemize}
To test the scalability of \tech{}, we also evaluate DYNAP-SE with 16 and 32 tiles. 

Table \ref{tab:hw_parameters} reports the hardware parameters of DYNAP-SE.

\vspace{-10pt}
\begin{table}[h!]
    \caption{Major simulation parameters extracted from \cite{dynapse}.}
	\label{tab:hw_parameters}
	\vspace{-10pt}
	\centering
	{\fontsize{6}{10}\selectfont
		\begin{tabular}{lp{5cm}}
			\hline
			Neuron technology & 65nm CMOS\\
			\hline
			Synapse technology & PCM\\
			\hline
			Supply voltage & 1.2V\\
			\hline
			Energy per spike & 50pJ at 30Hz spike frequency\\
			\hline
			Energy per routing & 147pJ\\
			\hline
			Switch bandwidth & 1.8G. Events/s\\
			\hline
	\end{tabular}}
	\vspace{-10pt}
\end{table}

\subsection{Evaluated Techniques}
\begin{figure}[h!]
	\centering
	\centerline{\includegraphics[width=0.45\columnwidth]{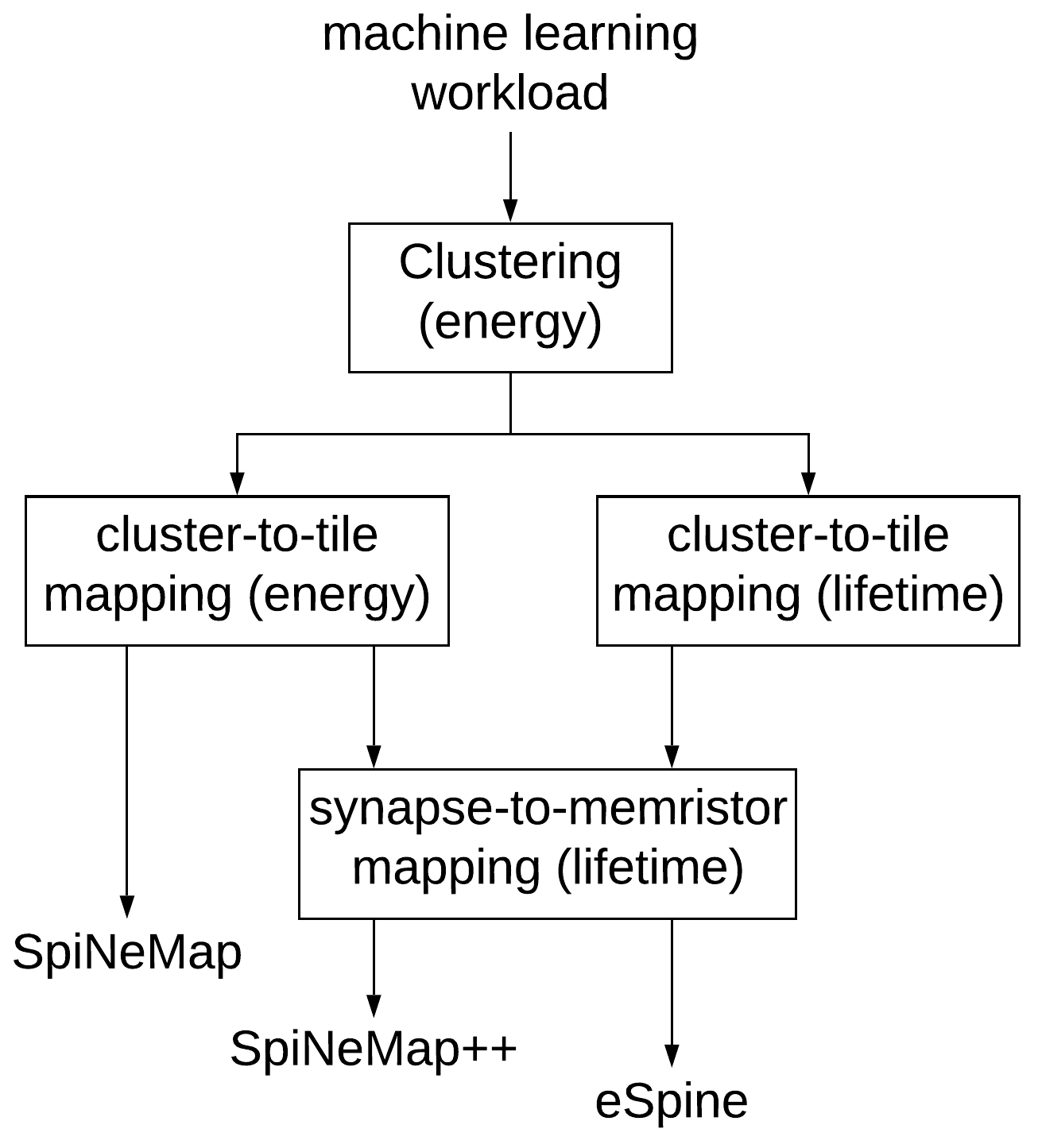}}
	\vspace{-10pt}
	\caption{Evaluated techniques.}
	\vspace{-10pt}
	\label{fig:sota}
\end{figure}

We evaluate the following techniques (see Fig.~\ref{fig:sota}).
\begin{itemize}
    \item \textbf{\prior{}:} This is the baseline technique to map SNNs to crossbars of a hardware. \prior{} generates clusters from an SNN workload, minimizing the inter-cluster communication. Clusters are mapped to tiles minimizing the energy consumption. Synapses of a cluster are implemented on memristors arbitrarily, without incorporating their endurance.
    \item \textbf{\priorpp{}:} \mr{This is an extension of \prior{}, where the cluster-to-tile mapping is performed using \prior{}, minimizing energy consumption, and the synapse-to-memristor mapping is performed using \tech{}, maximizing effective lifetime.}
    \item \textbf{\tech{}:} \mr{This is another extension of \prior{}. \tech{} uses only the clustering technique of \prior{}, thereby minimizing the inter-cluster communication, which also improves energy consumption and latency. The cluster-to-tile and synapse-to-memristor mappings are performed using PSO, maximizing the effective lifetime. Furthermore, \tech{} allows to explore the entire Pareto space of energy and lifetime.}
\end{itemize}
\vspace{-10pt}

\subsection{Evaluated Metric}
We evaluate the following metrics.
\begin{itemize}
    \item \textbf{Effective lifetime:} This is the minimum effective lifetime of all memristors in the hardware.
    \item \textbf{Energy consumption:} This is the total energy consumed on the hardware. We evaluate the static and dynamic energy as formulated in~\cite{shafik2015adaptive,das2016slowdown,das2013energy}.
    \item \textbf{Compilation time:} This is the time it takes for the PSO to find a solution.
\end{itemize}

\section{Results and Discussions}\label{sec:results}
\subsection{Normalized Lifetime}\label{sec:lifetime_results}
\begin{figure}[h!]
	\centering
	\vspace{-10pt}
	\centerline{\includegraphics[width=0.99\columnwidth]{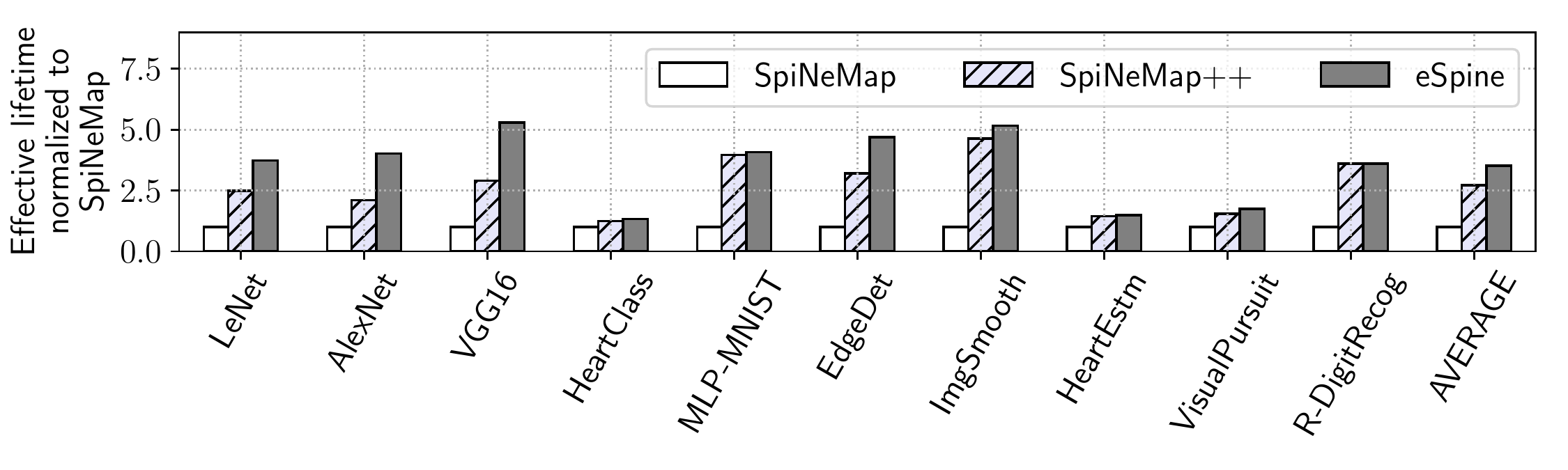}}
	\vspace{-10pt}
	\caption{Effective lifetime for the evaluated applications.}
	\vspace{-10pt}
	\label{fig:endurance_results}
\end{figure}

Figure \ref{fig:endurance_results} compares the effective lifetime obtained using each technique for each evaluated application on DYNAP-SE. We make the following two key observations.

First, between \prior{} and \priorpp{}, \priorpp{} has an average 2.7x higher effective lifetime than \prior{}. 
Although both \prior{} and \priorpp{} have
the same cluster-to-tile mapping, \priorpp{} maps synapses of a cluster intelligently on memristors of a crossbar, incorporating 1) the endurance variability of memristors in a crossbar and 2) the activation of synapses in a workload. Therefore, \priorpp{} has higher effective lifetime than \prior{}, which maps synapses arbitrarily to memristors of a crossbar.
Second, \tech{} has the highest effective lifetime than all evaluated techniques. The effective lifetime of \tech{} is higher than \prior{} and \priorpp{} by average 3.5x and 1.30x, respectively.  Although both \tech{} and \priorpp{} uses the same synapse-to-memristor mapping strategy, i.e., they both implement synapses with higher activation using memristors with higher endurance, the improvement of \tech{} is due to the PSO-based cluster-to-tile mapping, which maximizes the effective lifetime. Third, for some applications such as MLP-MNIST and R-DigitRecog, the effective lifetime using \tech{} is comparable to \priorpp{}. For these applications, the cluster-to-tile mapping of \prior{} is already optimal in terms of the effective lifetime. For other applications, \tech{} is able to find a better mapping, which improves the effective lifetime (by average 38\% compared to \priorpp{}).

\subsection{Energy Consumption}\label{sec:energy_results}
\begin{figure}[h!]
	\centering
	\vspace{-10pt}
	\centerline{\includegraphics[width=0.99\columnwidth]{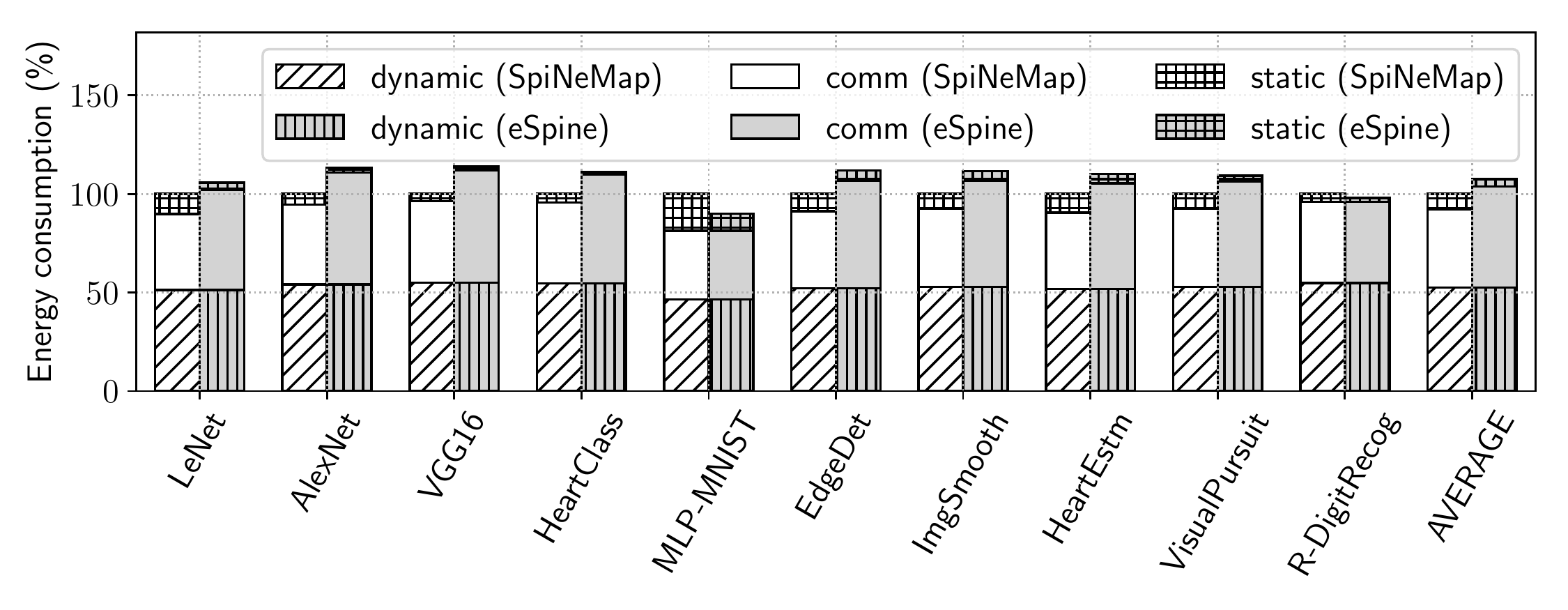}}
	\vspace{-10pt}
	\caption{Energy distribution for the evaluated applications.}
	\vspace{-10pt}
	\label{fig:energy_results}
\end{figure}

Figure \ref{fig:energy_results} reports the energy consumption of \prior{} and \tech{} on DYNAP-SE, distributed into 1) dynamic energy, which is consumed in crossbars to generate spikes (\texttt{dynamic}), 2) communication energy, which is consumed on the shared interconnect to communicate spikes between crossbars (\texttt{comm}), and 3) static energy, which is consumed in crossbars due to the leakage current through the access transistor of each memristor cell (\texttt{static}). We make the following four key observations.

First, the dynamic energy, communication energy, and static energy constitute respectively, 52.6\%, 39.4\%, and 8\% of the total energy consumption. Second, \tech{} does not alter spike generation, and therefore, the dynamic energy consumption of \tech{} is similar to \prior{}. Third, \tech{}'s cluster-to-tile mapping strategy is to optimize the effective lifetime, while \prior{} allocates clusters to tiles minimizing the energy consumption on the shared interconnect. Therefore, the communication energy of \prior{} is lower than \tech{} by an average of 21.4\%. Finally, \tech{} reduces the average temperature of each crossbar by implementing synapses with higher activation on longer current paths where memristors have lower self-heating temperature. Therefore, the leakage power consumption of \tech{} is on average 52\% lower than \prior{}.

\subsection{Energy Tradeoffs}\label{sec:energy_tradeoffs}
\begin{figure}[h!]
	\centering
	\vspace{-10pt}
	\centerline{\includegraphics[width=0.99\columnwidth]{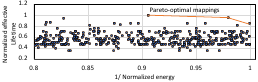}}
	\vspace{-10pt}
	\caption{Mapping explorations for LeNet.}
	\vspace{-10pt}
	\label{fig:pareto}
\end{figure}

Figure~\ref{fig:pareto} shows the normalized effective lifetime and the normalized energy of the mappings explored using the PSO algorithm for LeNet. 
The figure shows the mappings that are Pareto optimal with respect to lifetime and energy. 

\begin{figure}[h!]
	\centering
	\vspace{-10pt}
	\centerline{\includegraphics[width=0.99\columnwidth]{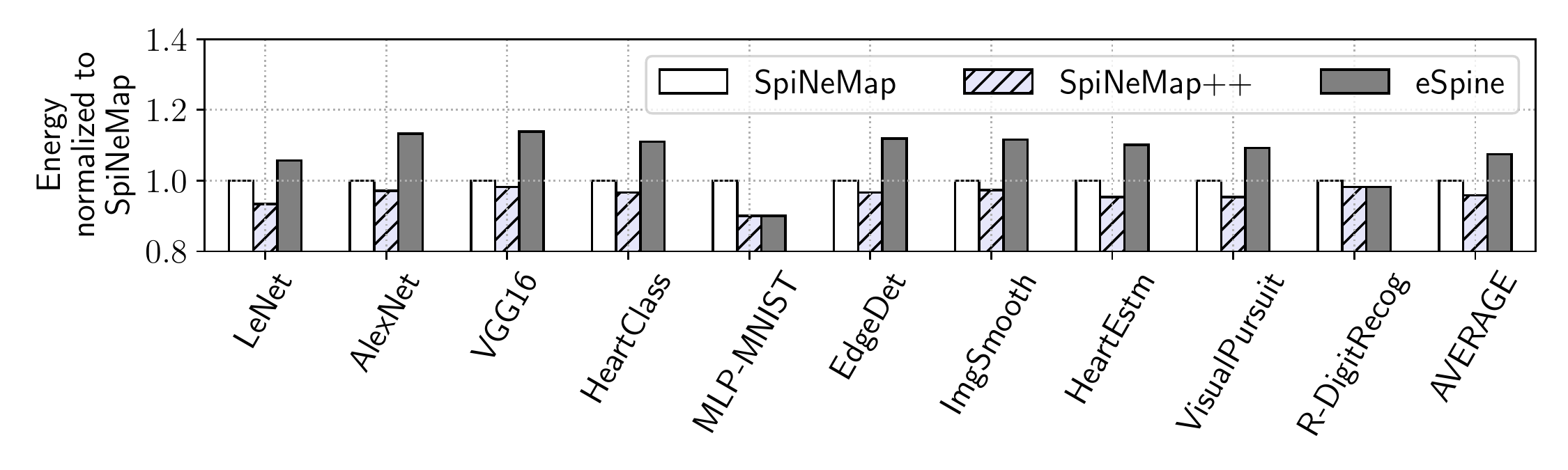}}
	\vspace{-10pt}
	\caption{Energy consumption for the evaluated applications.}
	\vspace{-10pt}
	\label{fig:energy_tradeoffs}
\end{figure}

Figure \ref{fig:energy_tradeoffs} reports the energy consumption of \prior{}, \priorpp{}, and \tech{} on DYNAP-SE for each evaluated application. We make the following two key observations.

\begin{figure*}[h!]%
    \centering
    \subfloat[Crossbar 1 (\prior{}).]{{\includegraphics[width=4.5cm]{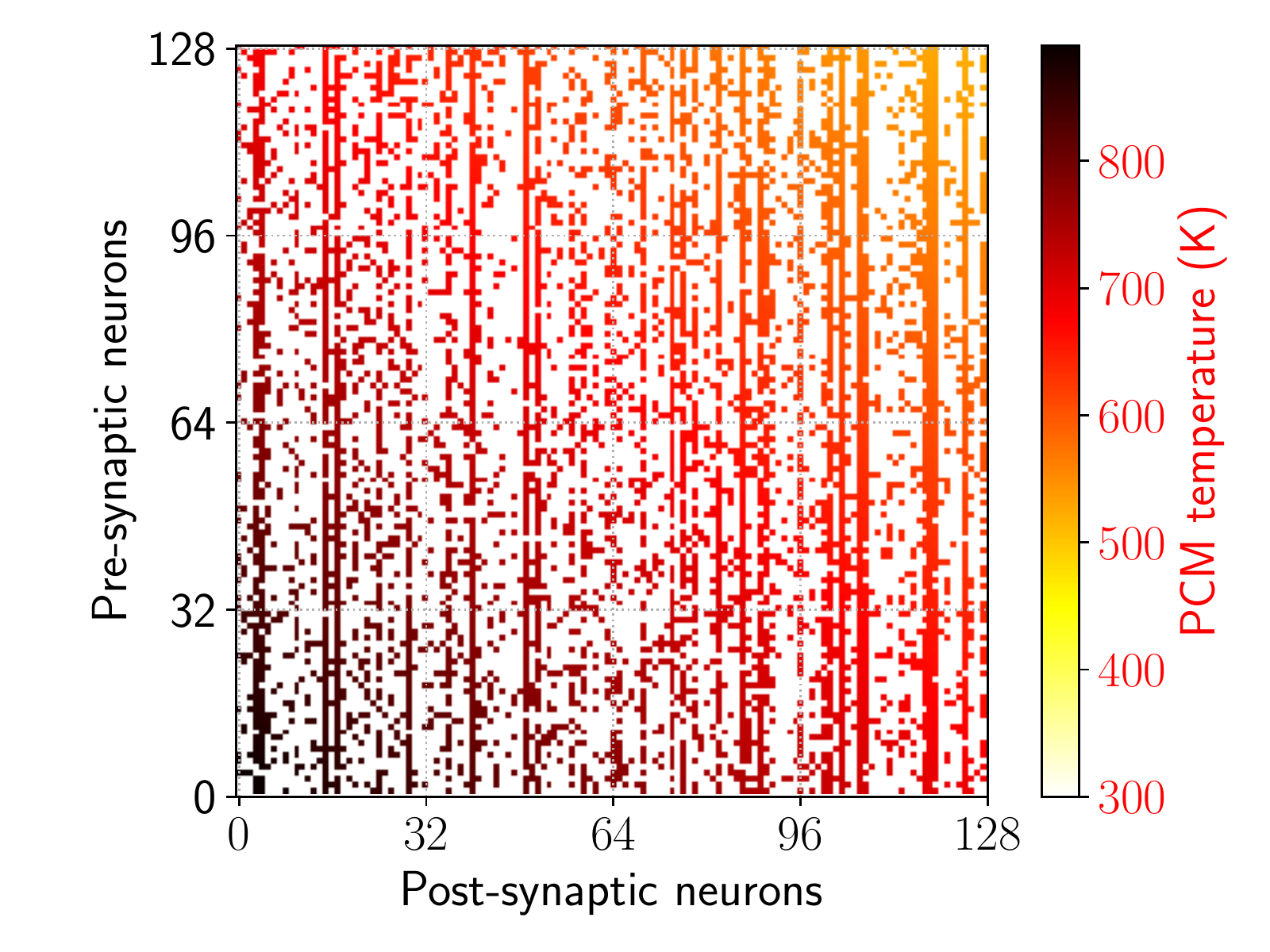} }}%
    \subfloat[Crossbar 2 (\prior{}).]{{\includegraphics[width=4.5cm]{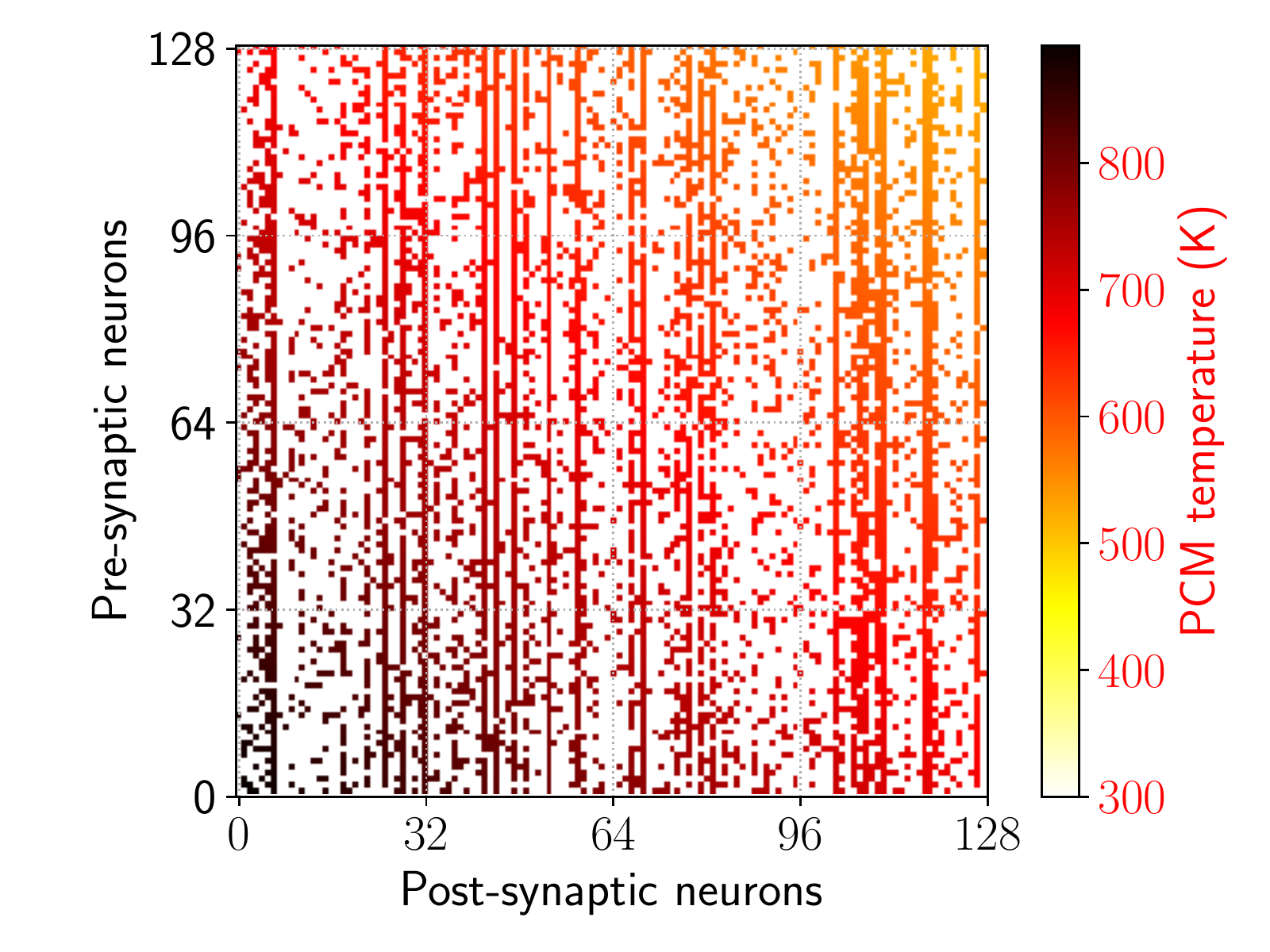} }}%
    \subfloat[Crossbar 3 (\prior{}).]{{\includegraphics[width=4.5cm]{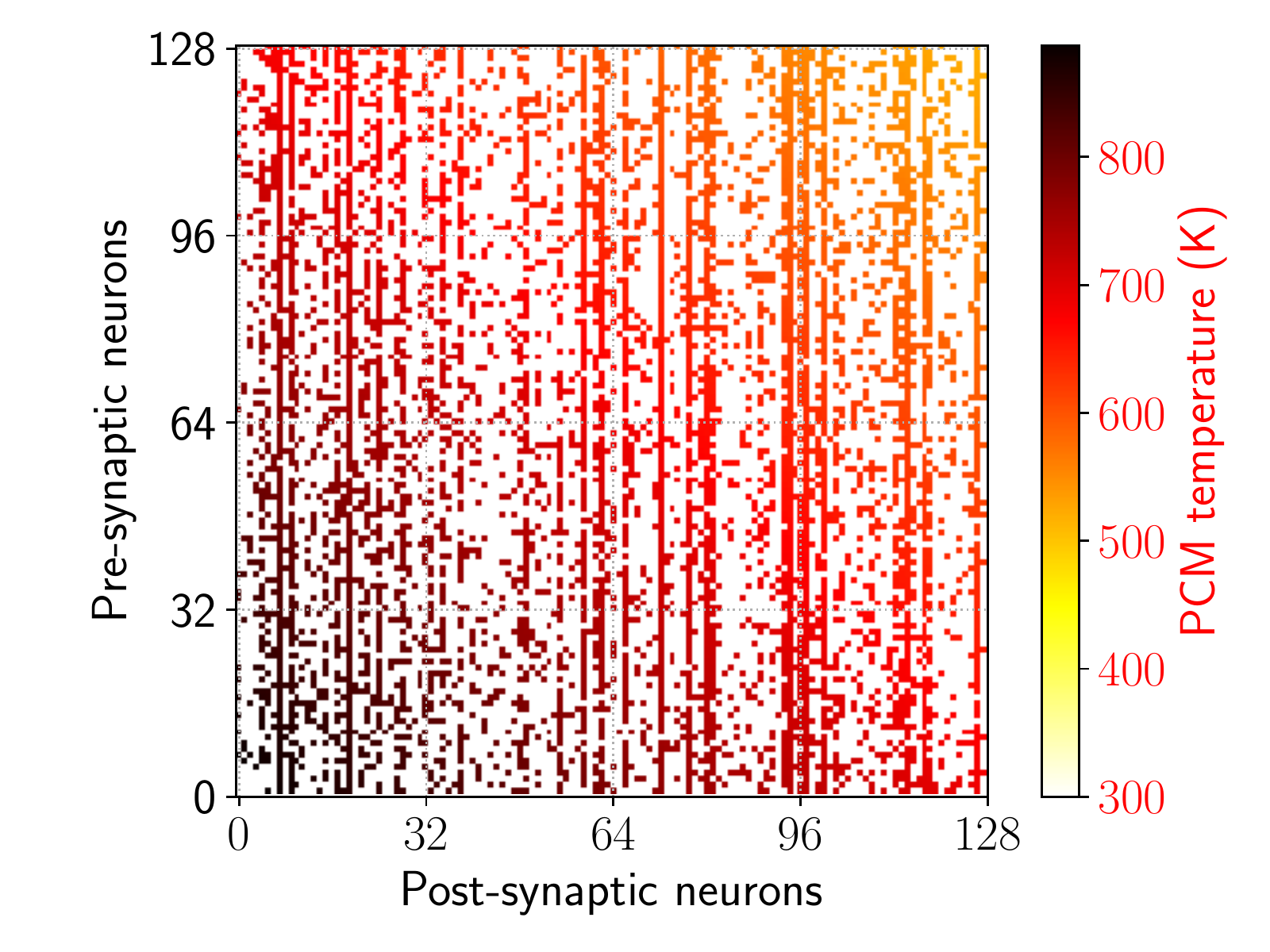} }}%
    \subfloat[Crossbar 4 (\prior{}).]{{\includegraphics[width=4.5cm]{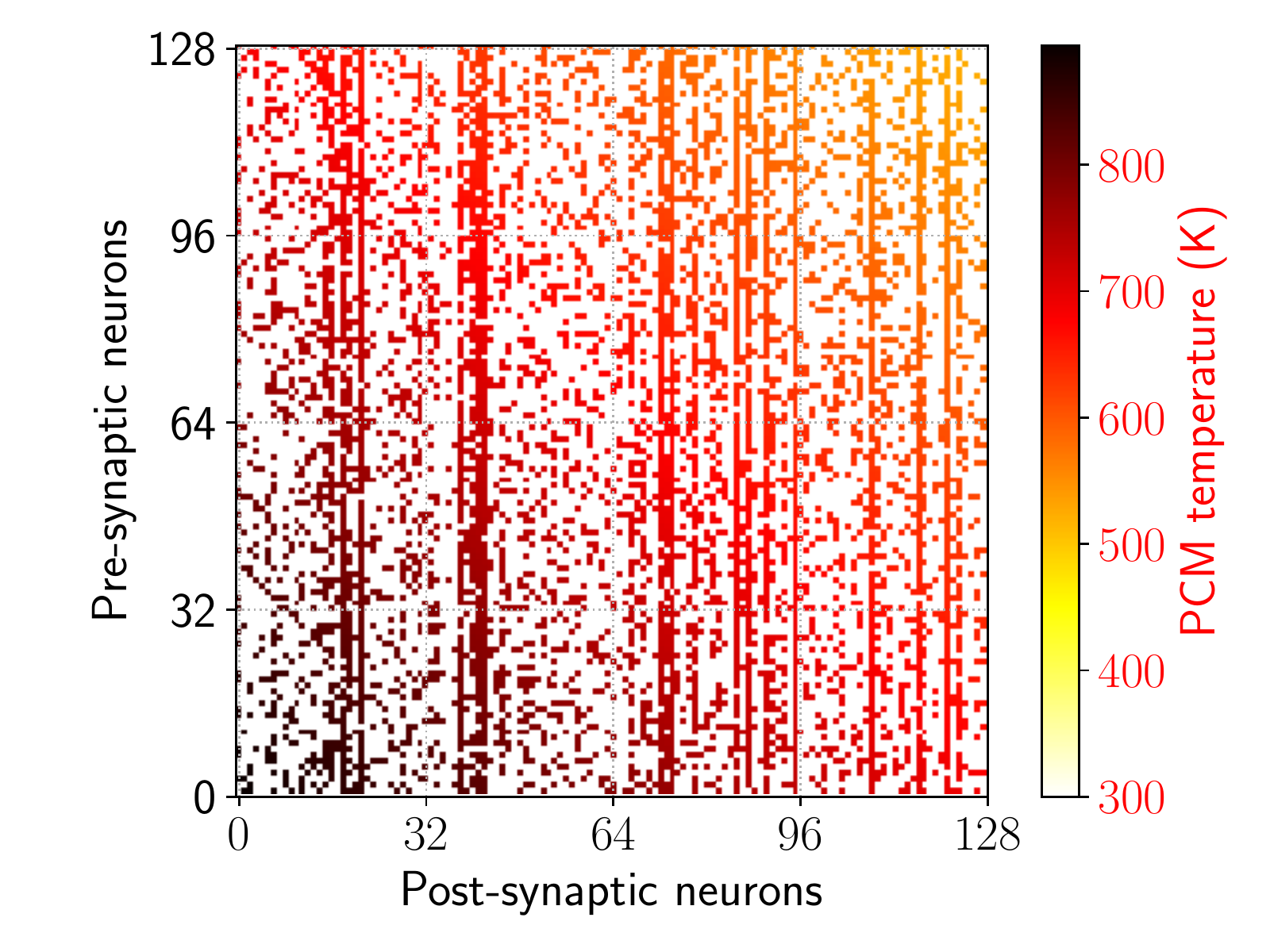} }}%
    \quad
    \subfloat[Crossbar 1 (\tech{}).]{{\includegraphics[width=4.5cm]{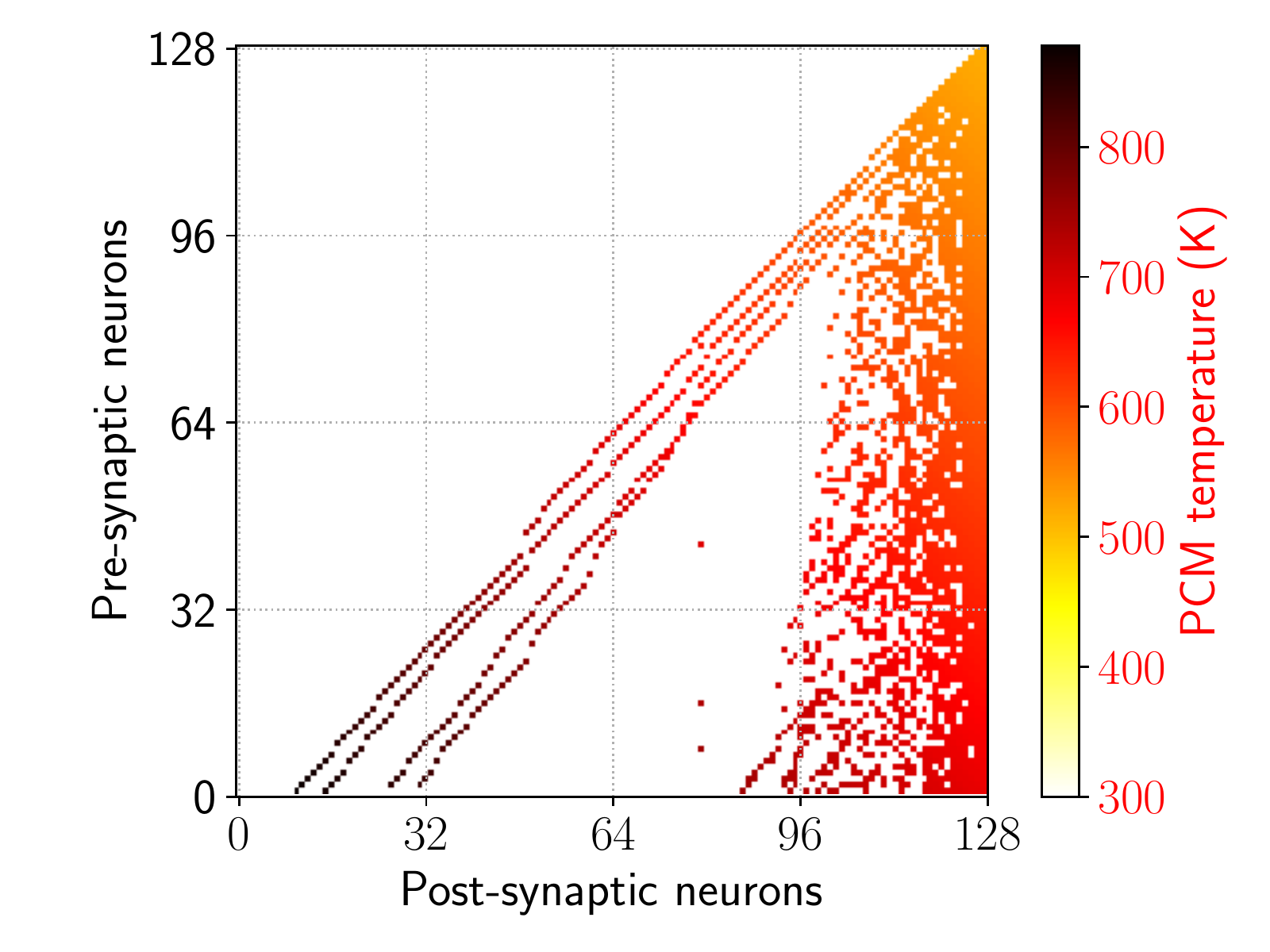} }}%
    \subfloat[Crossbar 2 (\tech{}).]{{\includegraphics[width=4.5cm]{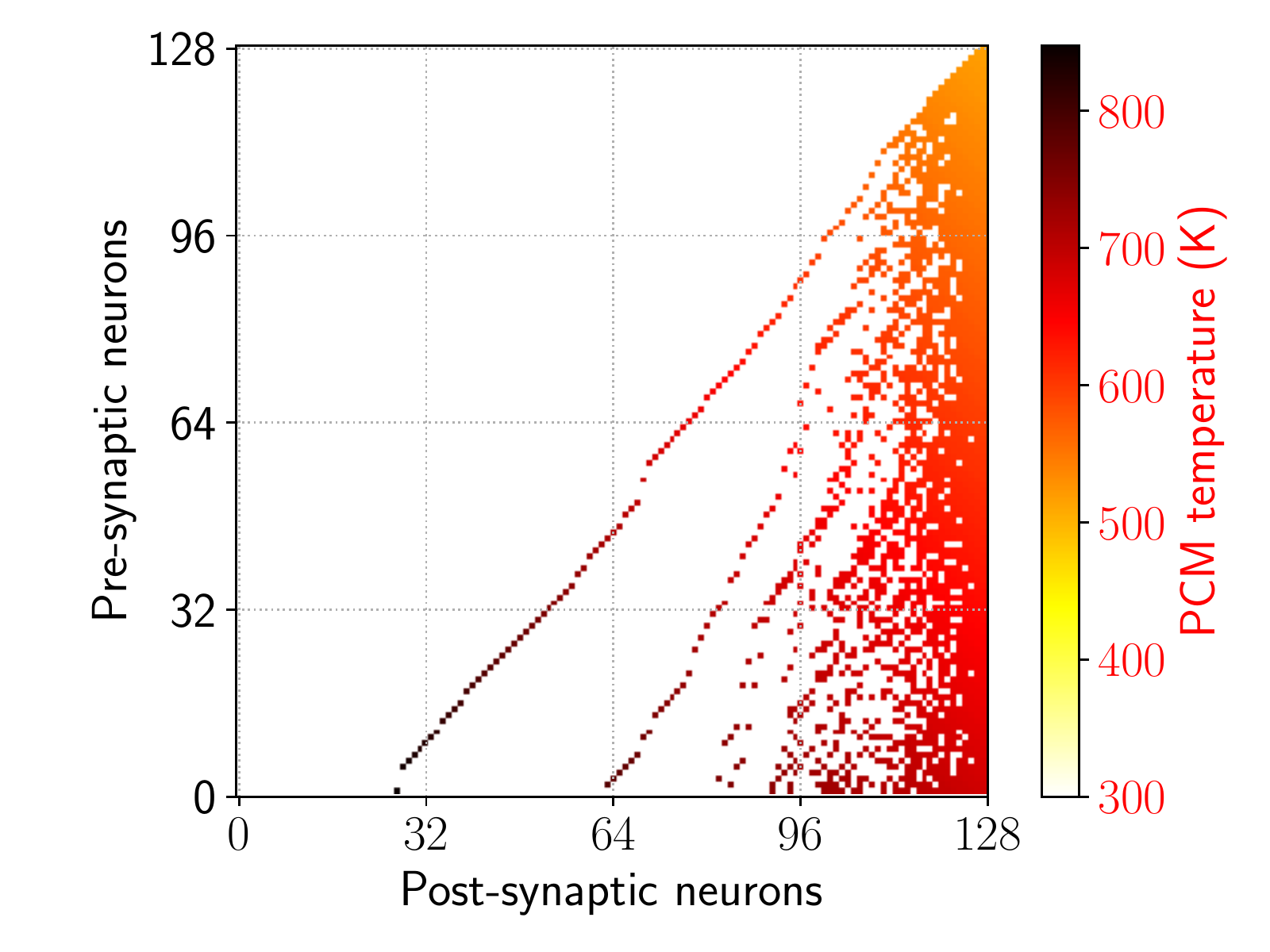} }}%
    \subfloat[Crossbar 3 (\tech{}).]{{\includegraphics[width=4.5cm]{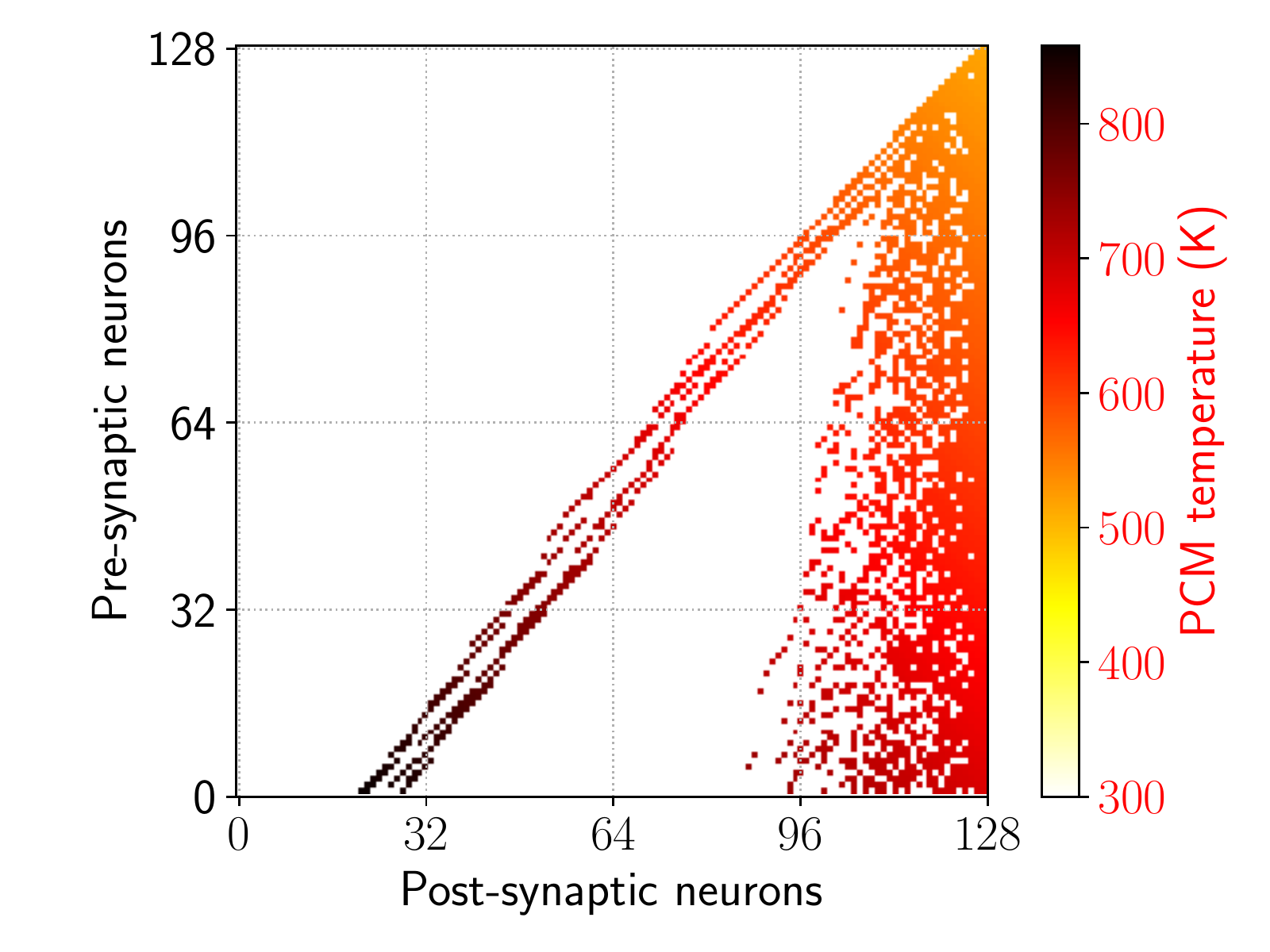} }}%
    \subfloat[Crossbar 4 (\tech{}).]{{\includegraphics[width=4.5cm]{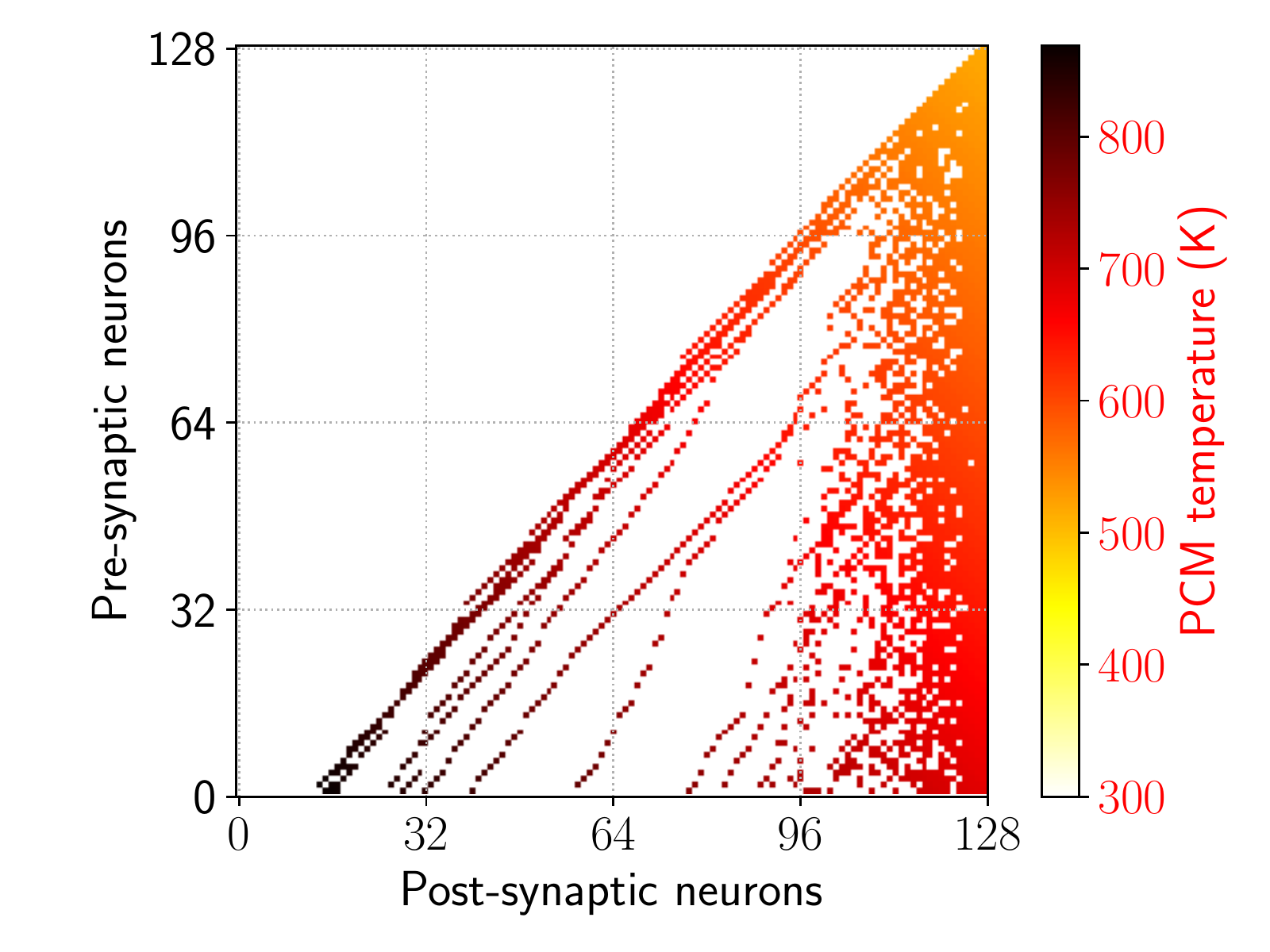} }}%
    \caption{Average temperature of the four crossbars in DYNAP-SE executing LeNet workload using \prior{} and \tech{}.}%
    \label{fig:lenet_results}%
\end{figure*}

First, the energy consumption of \priorpp{} is on average 4\% lower than \prior{}. This reduction is due to the reduction of leakage current, which is achieved by using memristors with lower self-heating temperature.
The energy consumption of \tech{} is higher than both \prior{} and \priorpp{} by an average of 7.5\% and 11.6\%, respectively. Although \tech{}, like \priorpp{}, lowers the static energy consumption by its intelligent synapse-to-memristor mapping, the higher energy consumption of \tech{} is due to the increase in the energy consumption on the shared interconnect of the hardware. However, by using an energy-aware clustering technique to begin with, \tech{} ensures that the overall energy consumption is not excessively high.
From the results of Sections~\ref{sec:lifetime_results} \& \ref{sec:energy_tradeoffs}, we make the following two key conclusions. 
First, \priorpp{}, which is \prior{} combined with the proposed synapse-to-memristor mapping, is best in terms of energy, achieving 2.7x higher lifetime than \prior{}. Second, \tech{}, which is our proposed cluster-to-tile and synapse-to-memristor mappings combined, is best in terms of lifetime, achieving 3.5x higher lifetime than \prior{}.

\subsection{\fix{Performance}}\label{sec:performance_results}
\fix{Table~\ref{tab:performance} reports the performance of the evaluated applications using \tech{} (Column 3). Results are compared against Baseline, which uses PyCARL~\cite{pycarl} to estimate the accuracy of these applications on hardware assuming that the current injected in each memristor is what is needed for its synaptic weight update (Column 2). The table also reports the accuracy using \tech{}, where the synaptic weights are scaled as proposed in~\cite{zhang2020lifetime} to compensate for the accuracy loss due to the current imbalance in a crossbar (Column 4). We make the following two key observations.}

\begin{table}[h!]
	\renewcommand{\arraystretch}{1.0}
	\setlength{\tabcolsep}{1.2pt}
	\caption{Accuracy of Baseline (PyCARL~\cite{pycarl}), \tech{}, and \tech{} combined with \cite{zhang2020lifetime} for the evaluated applications.}
	\label{tab:performance}
	\vspace{-10pt}
	\centering
	{\fontsize{6}{9}\selectfont
		\begin{tabular}{|c|c|c|c|c|c|c|c|}
			\hline
			\multirow{2}{*}{\textbf{Application}} & \multicolumn{3}{|c|}{\textbf{Accuracy (\%)}} & \multirow{2}{*}{\textbf{Application}} & \multicolumn{3}{|c|}{\textbf{Accuracy 
			(\%)}}\\
			\cline{2-4}\cline{6-8}
			& \textbf{Baseline} & \textbf{\tech{}} & \textbf{\tech{} + \cite{zhang2020lifetime}} & & \textbf{Baseline} & \textbf{\tech{}} & \textbf{\tech{} + \cite{zhang2020lifetime}}\\
            \hline
LeNet	&	85.1	&	84.2	&	85.0	&
AlexNet	&	90.7	&	88.7	&	89.8	\\
VGG16	&	69.8	&	64.4	&	67.8	&
HeartClass	&	63.7	&	59.2	&	62.4	\\
MLP-MNIST	&	91.6	&	91.3	&	91.6	&
EdgeDet	&	100	&	86	&	96.8	\\
ImgSmooth	&	100	&	100	&	100.0	&
HeartEstm	&	67.9	&	67.9	&	67.9	\\
VisualPursuit	&	47.3	&	47.3	&	47.3	&
R-DigitRecog	&	83.6	&	81.5	&	83.6	\\
            \hline
	    \end{tabular}
	 }
\end{table}

\fix{
First, the Baseline has the highest accuracy of all. This is because, the PyCARL framework of Baseline assumes that the current through all memristors in a crossbar are the same. Second, current imbalance can lead to a difference between the expected and actual synaptic plasticity based on the specific memristor being accessed. Therefore, we see an average 3\% reduction in accuracy using \tech{}.
However, the current imbalance-aware synapse update strategy, when combined with \tech{} can solve this problem. In fact, we estimate that the accuracy of machine learning applications using this synaptic update strategy is on average 2\% higher than \tech{} and only 1\% lower than the Baseline.
}

\subsection{Average Temperature}\label{sec:avg_temperature}
\mrr{Figure~\ref{fig:lenet_results} plots the average self-heating temperature of the PCM cells in four crossbars in DYNAP-SE executing LeNet workload using \prior{} and \tech{}. 
We make the following two observations.}

\mrr{
First, \tech{} maps active memristive synapses towards the top right corner of a crossbar. However, such mapping does not lead to a significant change in the ambient temperature. This is because of the the chalcogenide alloy (e.g., Ge${}_2$Sb${}_2$Te${}_5$ \cite{ovshinsky1968reversible}) used to build a PCM cell, which keeps the self-heating temperature of the cell concentrated at the interface between the heating element and the amorphous dome (see Figure~\ref{fig:pcm_memory_cell_integration}), with only a negligible spatial heat flow to the surrounding~\cite{pigot2018comprehensive}.
}

\mrr{
Second, the average self-heating 
temperature of \tech{} is lower than \prior{}. This is because of the synapse-to-memristor mapping technique of \tech{}, which places synapses with higher activation on longer current paths, where the self-heating temperature of a memristor is lower.
By reducing the average temperature, \tech{} lowers the leakage current through the access transistor of a memristor, which we discussed in Section~\ref{sec:energy_results}.
}

\subsection{Resource Scaling}\label{sec:resource_results}
\begin{figure}[h!]
	\centering
	\vspace{-10pt}
	\centerline{\includegraphics[width=0.99\columnwidth]{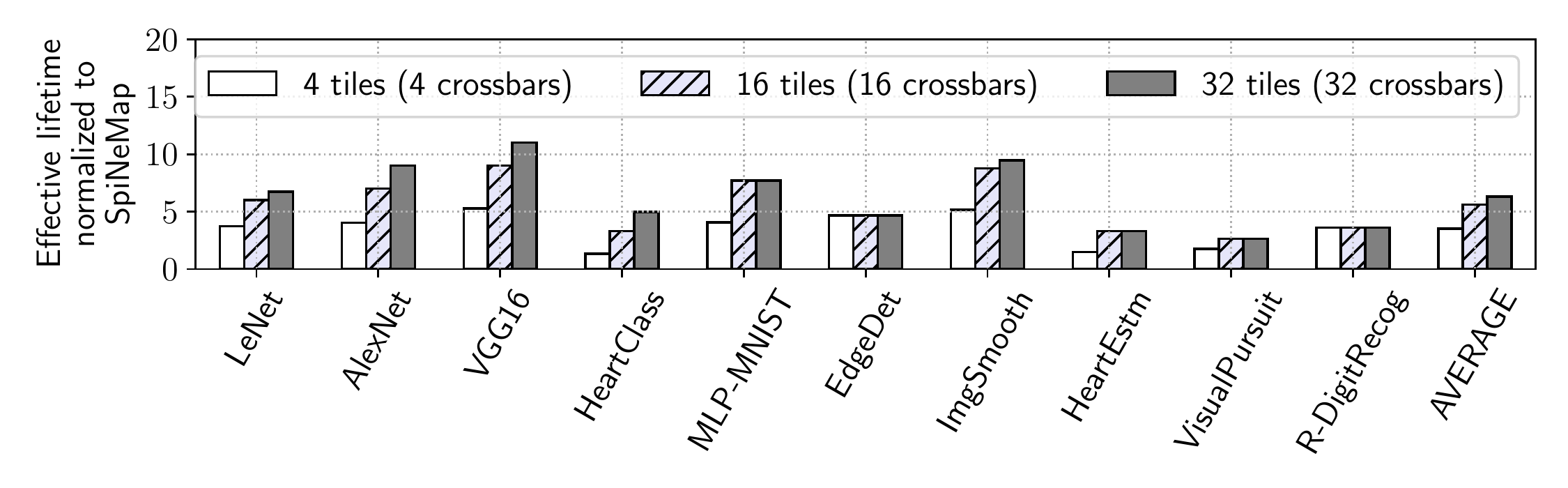}}
	\vspace{-10pt}
	\caption{Lifetime normalized to \prior{} for the evaluated applications on DYNAP-SE with 4, 16, and 32 tiles.}
	\vspace{-10pt}
	\label{fig:resource_results}
\end{figure}

Figure \ref{fig:resource_results} compares the lifetime normalized to \prior{} for each evaluated application on DYNAP-SE with 4-tile (4 crossbars), 16-tile (16 crossbars), and 32-tile (32 crossbars).

We observe that with 4, 16, and 32 tiles in the system, \tech{} provides an average 3.5x, 5.3x, and 6.4x lifetime improvement, respectively for the evaluated applications compared to \prior{}.
This is because with more tiles in the system, the workload gets distributed across the available crossbars of the hardware, resulting in lower average utilization of memristors, improving their lifetime.

\subsection{Compilation Time}\label{sec:solution_time_results}
Table~\ref{tab:compile_time} reports \tech{}'s compilation time and the effective lifetime normalized to \prior{} for three different settings of PSO iterations. We observe that as the number of PSO iterations is increased, the effective lifetime increases for all applications. This is because with increase in the number of iterations, the PSO is able to find a better solution. However, the compilation time also increases. \fix{We observe that the compilation time is significantly large for larger applications like VGG16 with 100 PSO iterations. However, we note that the PSO-based optimization is performed once at design-time. Furthermore, the PSO-iterations is a user-defined parameter, and therefore, it can be set to a lower value to generate a faster mapping solution, albeit a lower lifetime improvement.} 
Finally, we observe that increasing the PSO iterations beyond 100 leads to a significant increase in the compilation time for all applications with minimal improvement of their effective lifetime. 

\vspace{-10pt}
\begin{table}[h!]
	\renewcommand{\arraystretch}{1.0}
	\setlength{\tabcolsep}{0.8pt}
	\caption{Compilation time and solution quality tradeoff.}
	\label{tab:compile_time}
	\vspace{-10pt}
	\centering
	{\fontsize{6}{9}\selectfont
		\begin{tabular}{|r|c|c|c|c|c|c|}
			\hline
			\multirow{4}{*}{\textbf{Application}} & \multicolumn{2}{|c|}{\textbf{PSO Iterations = 1}} & \multicolumn{2}{|c|}{\textbf{PSO Iterations = 10}} & \multicolumn{2}{|c|}{\textbf{PSO Iterations = 100}}\\
			\cline{2-7}
			& \textbf{Compilation} & \textbf{Norm.} & \textbf{Compilation} & \textbf{Norm.} & \textbf{Compilation} & \textbf{Norm.}\\
			& \textbf{Time} & \textbf{Effective} & \textbf{Time} & \textbf{Effective} & \textbf{Time} & \textbf{Effective}\\
			& \textbf{(sec)} & \textbf{Lifetime} & \textbf{(sec)} & \textbf{Lifetime} & \textbf{(sec)} & \textbf{Lifetime}\\
			\hline
			LeNet	&	232.8	&	2.5	&	1,650.6	&	3.4	&	23,311.4	&	3.7	\\
AlexNet	&	331.7	&	2.1	&	2,431.8	&	3.1	&	45,617.4	&	4.0	\\
VGG16	&	886.8	&	2.9	&	8,156.0	&	4.2	&	110,123.6	&	5.3	\\
HeartClass	&	731.5	&	1.2	&	7,796.9	&	1.2	&	79,557.9	&	1.3	\\
MLP-MNIST	&	3.4	&	4.0	&	17.2	&	4.1	&	327.3	&	4.1	\\
EdgeDet	&	37.7	&	3.2	&	225.5	&	3.8	&	3,909.2	&	4.7	\\
ImgSmooth	&	26.2	&	4.6	&	91.1	&	4.6	&	1,327.4	&	5.2	\\
HeartEstm	&	109.0	&	1.4	&	595.1	&	1.4	&	7,303.6	&	1.5	\\
VisualPursuit	&	112.8	&	1.6	&	1,139.7	&	1.8	&	17,183.7	&	1.8	\\
R-DigitRecog	&	28.5	&	3.6	&	127.7	&	3.6	&	2,155.6	&	3.6	\\
    \hline
	\end{tabular}}
\end{table}

\section{Conclusion}\label{sec:conclusions}
In this work, we present \tech{}, a simple, yet powerful technique to improve the effective lifetime of memristor-based neuromorphic hardware in executing SNN-based machine learning workloads. \tech{} is based on detailed circuit simulations at different process, voltage, and temperature corners to estimate parasitic voltage drops on different current paths in a memristive crossbar. The circuit parameters are used in a compact endurance model to estimate the endurance variability in a crossbar. This endurance variability is then used within a design-space exploration framework for mapping neurons and synapses of a workload to crossbars of a hardware, ensuring that synapses with higher activation are implemented on memristors with higher endurance, and vice versa. The mapping is explored using an instance of the Particle Swarm Optimization (PSO). We evaluate \tech{} using 10 SNN workloads representing commonly-used machine learning approaches. Our results for DYNAP-SE, a state-of-the-art neuromorphic hardware demonstrate the significant improvement of effective lifetime of memristors in a neuromorphic hardware.


%

\appendices


\ifCLASSOPTIONcompsoc
  \section*{Acknowledgments}
\else
  \section*{Acknowledgment}
\fi
This work is supported by the National Science Foundation Faculty Early Career Development Award CCF-1942697 (CAREER: Facilitating Dependable Neuromorphic Computing: Vision, Architecture, and Impact on Programmability).

\ifCLASSOPTIONcaptionsoff
  \newpage
\fi

\bibliographystyle{IEEEtran}
\bibliography{commands,disco,external}

\begin{IEEEbiography}[{\includegraphics[width=1in,height=1.25in,clip,keepaspectratio]{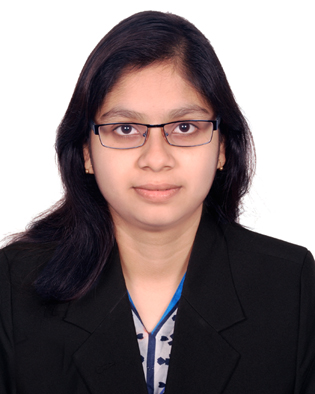}}]{Twisha Titirsha}
Twisha Titirsha is currently pursuing a Ph.D. degree from the Department of Electrical and Computer Engineering, Drexel University, Philadelphia. She received a Bachelor's degree from Military Institute of Science and Technology, Bangladesh in 2015. Her research interests include computer architecture, non-volatile memory and analog and/or mixed-signal circuit design.
\end{IEEEbiography}

\begin{IEEEbiography}[{\includegraphics[width=1in,height=1.25in,clip,keepaspectratio]{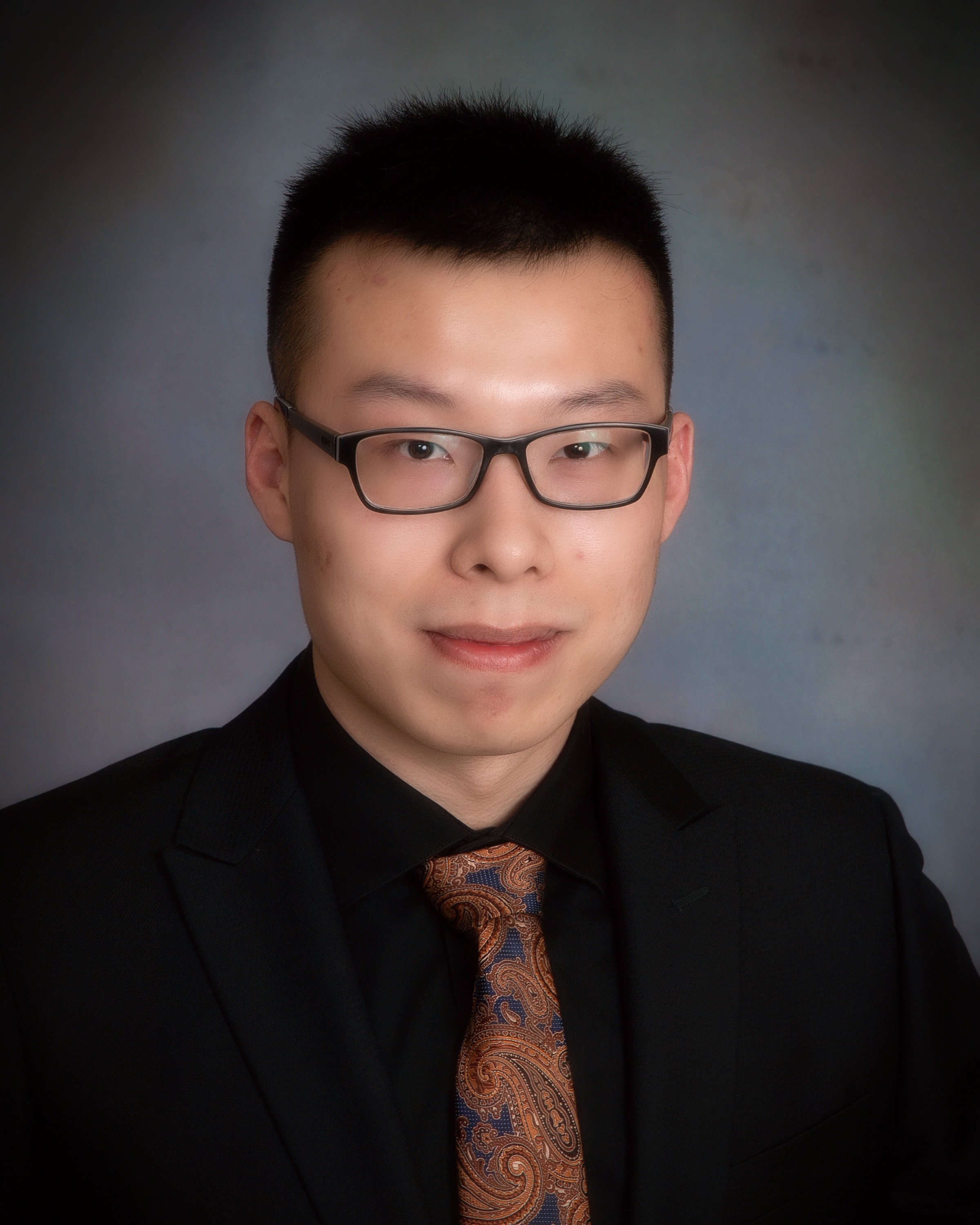}}]{Shihao Song}
Shihao Song is currently pursuing a Ph.D. degree from Drexel University under the supervision of Dr. Anup Das. He received a Bachelor's degree from Drexel University in 2017. His research interests include computer architecture, non-volatile memory, and compiler design for neuromorphic hardware and accelerators.
\end{IEEEbiography}



\begin{IEEEbiography}[{\includegraphics[width=1in,height=1.25in,clip,keepaspectratio]{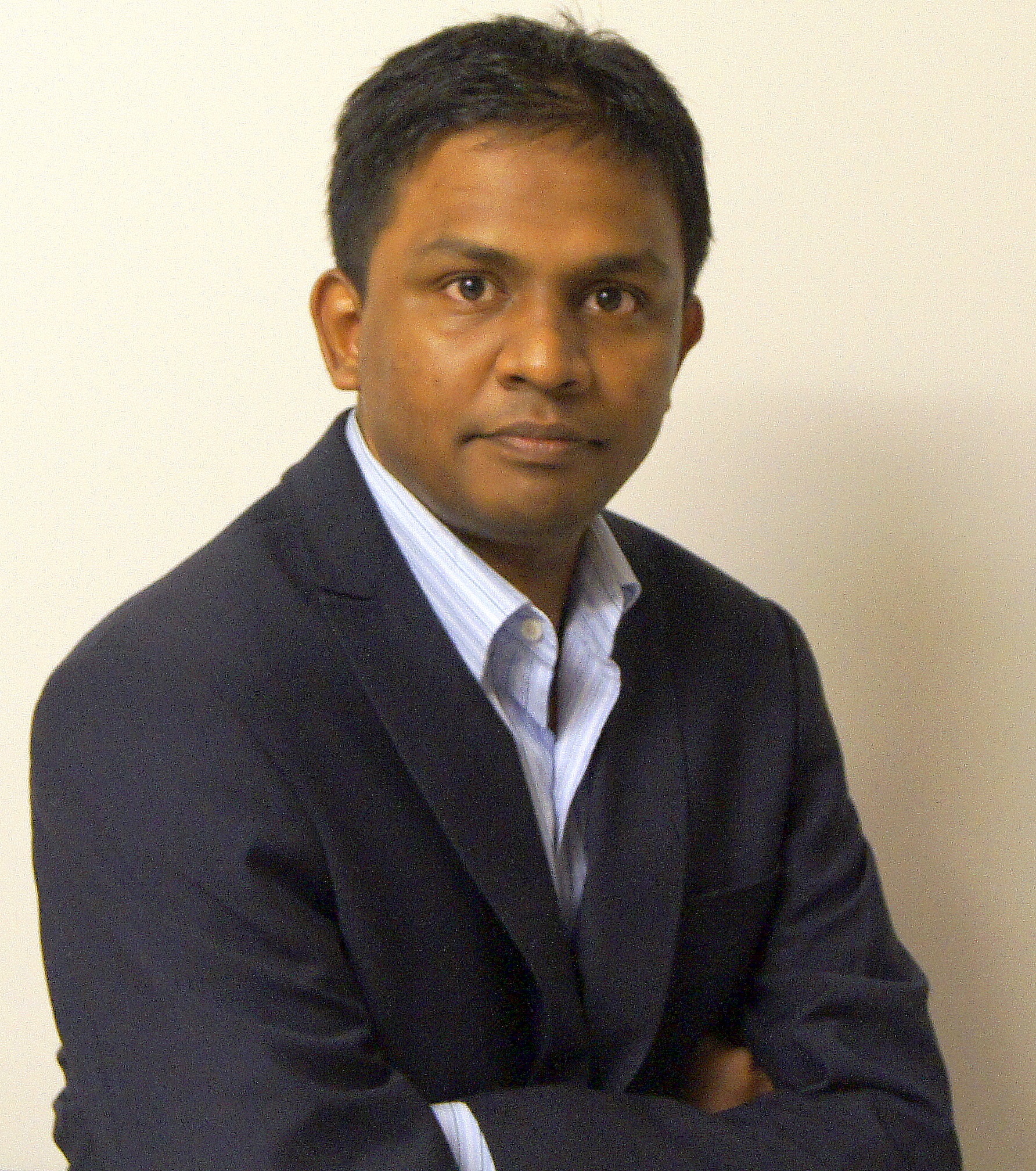}}]{Anup Das}
Dr. Anup Das is an Assistant Professor at Drexel University. He received a Ph.D. in Embedded Systems from National University of Singapore in 2014.  Following his Ph.D., he was a post-doctoral fellow at the University of Southampton and a researcher at IMEC. His research focuses on neuromorphic computing and architectural exploration. He is a senior member of the IEEE.
\end{IEEEbiography}

\begin{IEEEbiography}[{\includegraphics[width=1in,height=1.25in,clip,keepaspectratio]{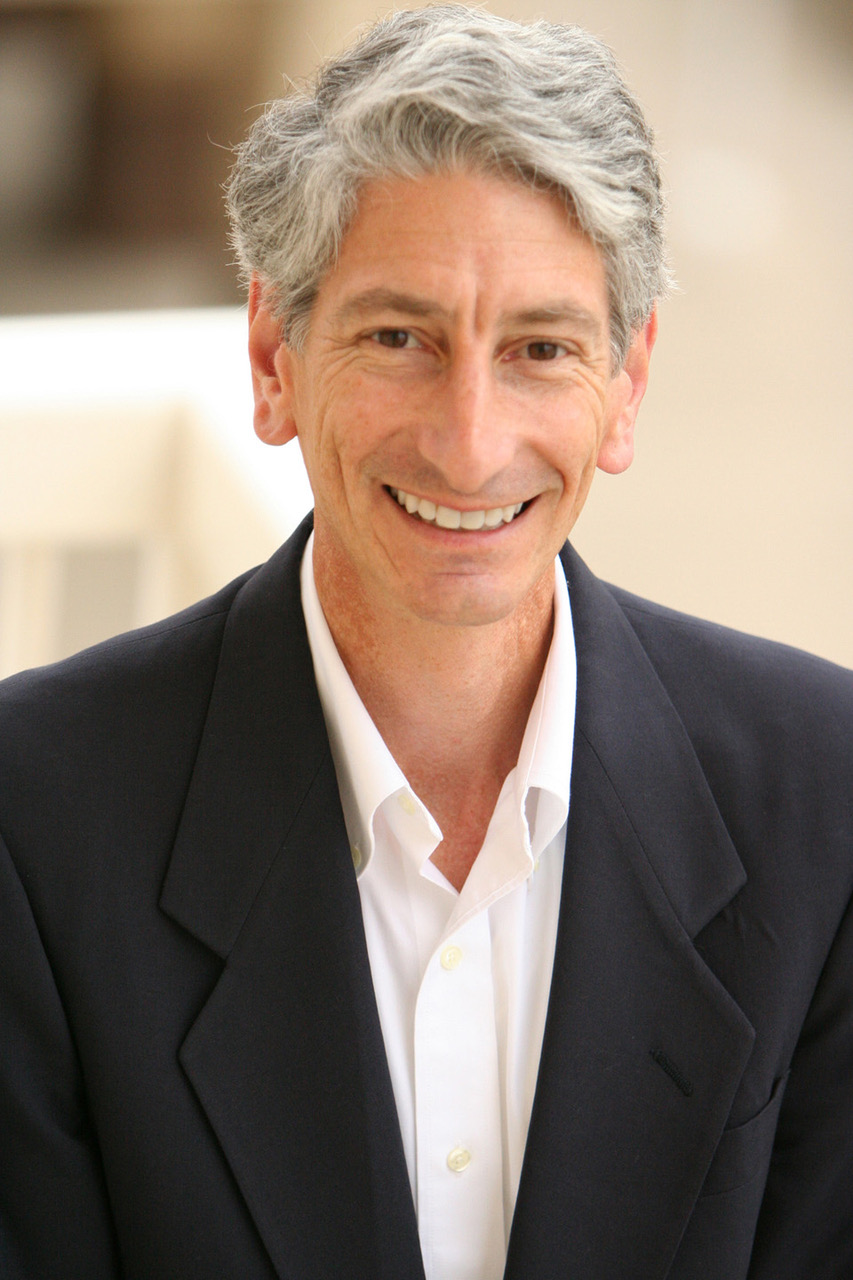}}]{Jeffrey L. Krichmar}
	Jeffrey L. Krichmar received a B.S. in Computer Science in 1983 from the University of Massachusetts at Amherst, a M.S. in Computer Science from The George Washington University in 1991, and a Ph.D. in Computational Sciences and Informatics from George Mason University in 1997. He spent 15 years as a software engineer on projects ranging from the PATRIOT Missile System at the Raytheon Corporation to Air Traffic Control for the Federal Systems Division of IBM. From 1999 to 2007, he was a Senior Fellow in Theoretical Neurobiology at The Neurosciences Institute. He currently is a professor in the Department of Cognitive Sciences and the Department of Computer Science at the University of California, Irvine. 
	He is a Senior Member of IEEE and the Society for Neuroscience.
\end{IEEEbiography}

\begin{IEEEbiography}[{\includegraphics[width=1in,height=1.25in,clip,keepaspectratio]{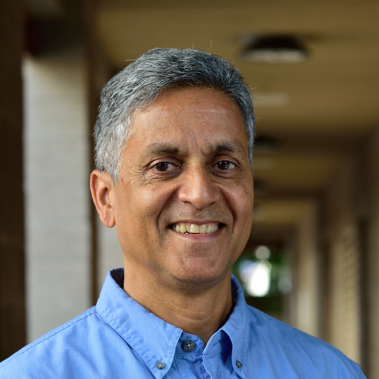}}]{Nikil D. Dutt}
	Nikil D. Dutt (F) received a Ph.D. in Computer Science from the University of Illinois at Urbana-Champaign in 1989, and is currently a Distinguished Professor of
Computer Science, Cognitive Sciences, and EECS at the University of California, Irvine. He is also a Distinguished Visiting Professor in the CSE department at IIT Bombay, India. Dutt’s research interests are in embedded systems, electronic design automation (EDA), computer systems architecture and software, healthcare IoT, and brain-inspired architectures and computing. 
	He is a Fellow of the ACM, Fellow of the IEEE, and recipient of the IFIP Silver Core Award.
\end{IEEEbiography}

\begin{IEEEbiography}[{\includegraphics[width=1in,height=1.25in,clip,keepaspectratio]{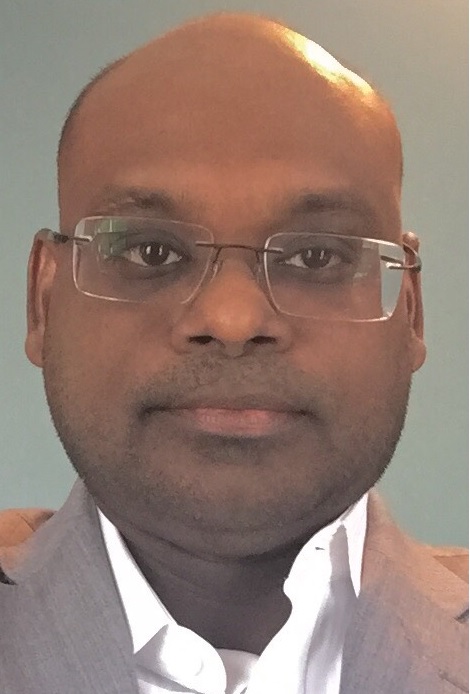}}]{Nagarajan Kandasamy}
	Nagarajan Kandasamy is a Professor in the Department of Electrical and Computer Engineering at Drexel University. His current research interests are in the areas of computer architecture, parallel processing, and embedded and real-time systems. His research has been funded by the National Science Foundation, the U.S. Army Research Office, and the Office of Naval Research, among others. He is a recipient of the NSF CAREER award; best student paper awards in the IEEE Conference on Autonomic Computing, and the Pacific Rim Dependable Computing conference, for work related to performance management and fault detection in computing systems; and a featured article in the Physics in Medicine and Biology Journal for work on GPU-accelerated algorithms for medical imaging. Dr. Kandasamy received his Ph.D. in Computer Science and Engineering from the University of Michigan. He is a senior member of the IEEE. 
\end{IEEEbiography}

\begin{IEEEbiography}[{\includegraphics[width=1in,height=1.25in,clip,keepaspectratio]{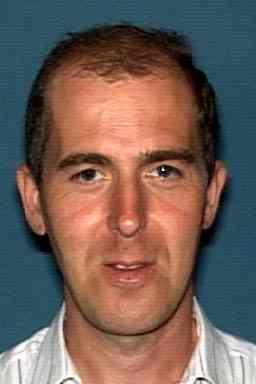}}]{Francky Catthoor}
	Dr. Francky Catthoor received  a Ph.D. in EE from the Katholieke Univ. Leuven,
	Belgium in 1987.  Between 1987 and 2000,  he  has headed  several research
	domains in the area of synthesis techniques and architectural methodologies.
	Since 2000 he is  strongly involved in other activities at IMEC including
	deep submicron technology aspects, IoT and biomedical platforms, and
	smart photovoltaic modules, all at IMEC Leuven,  Belgium.   Currently he is
	an IMEC fellow.
	He is also part-time full professor at the EE department of the KULeuven.
	He has been associate editor for several IEEE and ACM journals.
	He was elected IEEE fellow in 2005.
\end{IEEEbiography}




\end{document}